\documentclass[preprint,review,12pt]{elsarticle}

\usepackage{graphicx}
\usepackage{amssymb}
\usepackage{amsmath}
\usepackage{amsthm}
\usepackage{amsfonts,latexsym}
\usepackage{epsfig}
\usepackage{algorithm}
\usepackage{algpseudocode}
\usepackage{subcaption}
\usepackage[dvipsnames]{xcolor}
\usepackage{tikz}
\usepackage{amsfonts}
\usepackage{bm}
\usepackage{tikz}
\DeclareMathOperator*{\argmin}{\arg\!\min}
\newcommand{\norm}[1]{\| #1 \|}

\DeclareMathOperator*{\sign}{sgn}
\usepackage{multirow}
\usepackage{glossaries}
\usepackage[margin=1in]{geometry}
\makeglossaries

\newglossaryentry{Extended SEIRS}
{
        name=Extended SEIRS,
        description={Extended SEIRS model based on neutron transport}
}

\newglossaryentry{$f^{BDLSTM}$}
{
        name=$f^{BDLSTM}$,
        description={Reduced order model of the extended SEIRS model using a bidirectional LSTM}
}

\newglossaryentry{BLUE}
{
        name=BLUE,
        description={Best Linear Unbiased Estimator}
}

\newglossaryentry{$f^{BDLSTM+BLUE}$}
{
        name=$f^{BDLSTM+BLUE}$,
        description={Reduced order model of the extended SEIRS model using a bidirectional LSTM with the Best Linear Unbiased Estimator}
}

\newglossaryentry{$f^{PredictiveGAN}$}
{
        name=$f^{PredictiveGAN}$,
        description={Reduced order model of the extended SEIRS model using a predictive Generative Adversarial Network}
}

\usepackage{hyperref}
\usepackage{subcaption}
\usepackage{lineno}

\journal{Neurocomputing}

\begin{document}

\begin{frontmatter}


\title{Digital twins based on bidirectional LSTM and GAN for modelling the COVID-19 pandemic}



\author[a]{C\'esar Quilodr\'an-Casas\corref{cor1}}
    \ead{cesar.quilodran-casas13@imperial.ac.uk}
    \cortext[cor1]{Corresponding author}
\author[b]{Vinicius L. S. Silva}
\author[a]{Rossella Arcucci}
\author[b]{Claire E. Heaney}
\author[a]{YiKe Guo}
\author[a,b]{Christopher C. Pain}

\address[a]{Data Science Institute, Department of Computing, Imperial College London, UK}
\address[b]{Department of Earth Science $\&$ Engineering, Imperial College London, UK}

\begin{abstract}
The outbreak of the coronavirus disease 2019 (COVID-19) has now spread throughout the globe infecting over 150 million people and causing the death of over 3.2 million people. Thus, there is an urgent need to study the dynamics of epidemiological models to gain a better understanding of how such diseases spread. While epidemiological models can be computationally expensive, recent advances in machine learning techniques have given rise to neural networks with the ability to learn and predict complex dynamics at reduced computational costs.
Here we introduce two digital twins of a SEIRS model applied to an idealised town. The SEIRS model has been modified to take account of spatial variation and, where possible, the model parameters are based on official virus spreading data from the UK. We compare predictions from a data-corrected Bidirectional Long Short-Term Memory network and a predictive Generative Adversarial Network. The predictions given by these two frameworks are accurate when compared to the original SEIRS model data.

Additionally, these frameworks are data-agnostic and could be applied to towns, idealised or real, in the UK or in other countries. Also, more compartments could be included in the SEIRS model, in order to study more realistic epidemiological behaviour.
\end{abstract}


\end{frontmatter}


\section{Introduction}

The outbreak of the coronavirus disease 2019 (COVID-19) has now spread throughout the world, infecting over 153 million reported individuals as of May 4th 2021 \citep{dong2020interactive}. Globally, at least 3.2 million deaths have been directly attributed to COVID-19 \citep{dong2020interactive} and this number continues to rise. There is a lack of information and uncertainty about the dynamics of this outbreak, thus, there is an urgent need for research in this field to help with the mitigation of this pandemic \citep{park2020systematic}. Agent-based models \citep{auchincloss2008new, cuevas2020agent, shamil2021agent} and SEIR-type models \citep{li1995global, Radulescu2020} have been widely used to study epidemiological problems. However, when modelling complex scenarios, these models can become computationally expensive, e.g.~such models may have many millions of degrees of freedom that must be solved at every time step \citep{basu2013complexity, rock2014dynamics}. Also, the time steps may be small to resolve the transport of people around a domain. For instance, in a model of a town, a person in a car or train may travel large distances in just a few minutes \citep{cameron2004trends}. This advection can have limitations in terms of Courant number restrictions~\cite{pavlidis:14} based on the spatial resolution, as well as the speed of the transport. Furthermore, these models may have a set of variables for each member of a population. Thus, if a country is modelled with many millions of people, the computational expense of such models becomes an issue and they may even become intractable \citep{eubank2002scalable}. This has motivated the current research on accurate surrogates or reduced-order Models (ROMs) for virus modelling. Although ROMs have been developed in fields such as fluid dynamics, they are new for virus modelling. For this new application area, we study a simple test case to try to understand the application of these methods to virus modelling. The prize of an accurate and fast ROM means that it may be readily used, possibly interactively, to explore different control measures, to assimilate data into the models, and to help determine the spatial and future temporal variation of infections.  We may need to develop new ROM approaches to meet the demands of this new virus application area and explore the relative merits of existing and new ROM approaches which is the focus of this paper.



In this paper, we compare two methods for creating a digital twin of a SEIRS model \citep{cooke1996analysis, song2019spatial} that has been modified to take account of spatial variation. These digital twins or non-intrusive reduced-order models (NIROMs) are used to approximate future states of the model which are compared against the ground truth. The first experiment uses a data-corrected (via optimal interpolation) Bidirectional Long Short-term memory network (BDLSTM), while the second experiment utilises a Generative Adversarial Network (GAN).

NIROMs have been used with success in several fields, to speed up computational models without losing the resolution of the original model~\citep{xiao2015non, hesthaven2018non, quilodran2018fast, casas2020reduced} and without the need to make changes to the code of the high-fidelity model. Typically, the first stage in constructing a NIROM is to reduce the dimension of the problem by using compression methods such as Principal Component Analysis (PCA)~\citep{lever2017points}, autoencoders, or a combination of both \citep{casas2020urban, quilodran2021adversarial, phillips2020autoencoder}. Solutions from the original computational model (known as snapshots) are then projected onto the lower-dimensional space, and the resulting snapshot coefficients are interpolated in some way, to approximate the evolution of the model. This interpolation, which approximates unseen states of the model, constitutes the second stage of the NIROM. Originally, classical interpolation methods were used, such as cubic interpolation~\cite{Bui-Thanh2003}, radial basis functions~\cite{Benamara2017, Xiao2019} and Kriging~\cite{Aversano2019}. Recently, non-intrusive reduced-order methods (sometimes referred to as model identification methods~\cite{Kaiser2014, Wang2018} or described by the more general term of digital twins~\cite{Rasheed2020, Moya2020, Kapteyn2020}) have taken advantage of machine learning techniques, using multi-layer perceptrons~\citep{hesthaven2018non}, cluster analysis~\citep{Kaiser2014},  LSTMs~\citep{Wang2018, Ahmed2019, Kherad2020, quilodran2021adversarially} and Gaussian Process Regression~\citep{Guo2018}. In this work, we use PCA (also known as Proper Orthogonal Decomposition) to reduce the dimension of the original system, and for the interpolation or prediction, we compare a data-corrected BDLSTM with a predictive GAN. The LSTM network, originally described in \citep{hochreiter1997long}, is a special kind of recurrent neural network (RNN) that is stable, powerful enough to be able to model long-range time dependencies \citep{xingjian2015convolutional} and overcomes the vanishing gradient problem \citep{greff2016lstm}. A further development was made to this network inspired by bidirectional RNNs \citep{schuster1997bidirectional}, in which sequences of data are processed in both forward and backward directions. The resulting BDLSTMs have been proven to be better \citep{cui2018deep} than unidirectional ones, as the former can capture the
forward and backward temporal dependencies in spatiotemporal data \citep{cui2020stacked}, in many fields such as speech recognition \citep{graves2013hybrid} and traffic control \citep{cui2018deep}. Bidirectional LSTMs have also been used in text classification \citep{liu2019bidirectional}, predicting efficient remaining useful life of a system \citep{elsheikh2019bidirectional}, and urban air pollution forecasts \citep{quilodran2021adversarially}. LSTMs are widely recognised as one of the most effective sequential models \citep{goodfellow2016deep} for times series predictions. We compare the performance of LSTMs with GANs \citep{goodfellow:14} which are known for retaining realism. GANs have shown impressive performance for photo-realistic high-quality images of faces \citep{karras2019style, karras2020analyzing}; image to image translation \citep{chu2017cyclegan}; synthetical medical augmentation \citep{frid2018gan}; cartoon image generation \citep{liu2018auto}, amongst others. The basic idea of GANs is to simultaneously train a discriminator and a generator, where the discriminator aims to distinguish between real samples and generated samples; while the generator tries to fool the discriminator by creating fake samples that are as realistic as possible. The GAN is a generative model and its use in making predictions in time is a recent development~\cite{Silva2020}. By learning a distribution that fits the training data, the aim is that new samples, taken from the learned distribution formed by the generator, will remain `realistic' over time and will not diverge.

Although non-intrusive reduced-order modelling has not been applied to epidemiological problems, as far as we are aware, neural networks have been used to model the spread of viruses. Previous studies have used Long Short-term Memory networks for COVID-19 predictions: Modified SEIR predictions of the trend of the epidemic in China \citep{yang2020modified}, general outbreak prediction with machine learning \citep{ardabili2020covid}, time series forecasting of COVID-19 transmission in Canada \citep{chimmula2020time}, and predicting COVID-19 incidence in Iran \citep{ayyoubzadeh2020predicting}, amongst others. Generative networks have also been used to model aspects of the COVID-19 outbreak, mainly used in image recognition, e.g.~chest X-rays \citep{khalifa2020detection, wang2020covid}. Bayesian updating has also been applied to COVID-19 by \citet{wang2020bayesian}. Furthermore, \citep{bao2020covid} used spatio-temporal conditional GANs for estimating the human mobility response to COVID-19.

Whilst LSTM and GAN have been used to study COVID-19, the datasets employed in the aforementioned studies differ from our approach since 1) our approach uses a spatio-temporal dataset, rather than scalars evolving in time; 2) the dimensionality of the spatio-temporal model is reduced during the compression stage of NIROM. The NIROM transforms the spatio-temporal problem into a multivariate time-series problem. The advantages of using an LSTM and GAN are that the former is an effective sequential model \citep{goodfellow2016deep} and the latter can learn the underlying data distribution, reducing the forecast divergence \citep{yoon2019time}. However, the disadvantages of these models include the divergence that can occur in results generated by LSTMs after a certain time \citep{breuel2015benchmarking}, hence the data-correction, and, for GANs, the expense incurred during training and the difficulty of converging given their adversarial nature \citep{goodfellow:14}.

The novelty of this paper lies in the use of data-corrected forecasts with the state-of-the-art LSTM, and a comparison between a digital twin based on this, and one based on GAN methods that are novel for prediction in time. In summary, the main novelties and contributions of this paper are:
\begin{itemize}
    \item The application of reduced-order modelling to virus/epidemiology modelling.
    \item The application of the novel data-corrected BDLSTM-based ROM approach. This is the first time that the data-corrected BDLSTM has been incorporated within a ROM. Using data from the SEIRS model solution, optimal interpolation is included in the prediction-correction cycle of the BDLSTM to stabilise the forecast and to achieve improved accuracy.
    \item Comparison is made between time-series predictions of two digital twins: one based on the state-of-the-art LSTM and the other based on a GAN, a recent network that is known for its realistic predictions. The GAN can generate time sequences from random noise that are constrained to generate a forecast.
\end{itemize}



The structure of this paper is as follows. Section~\ref{section:seir} introduces the classical SEIRS model and the extended SEIRS model, which takes account of spatial variation. The SEIRS model in this paper also includes an additional way of categorising people according to their environment. Section~\ref{section:Methods} presents the methodology of the two digital twins (based on results from the extended SEIRS model) and explains how the predictions are performed. The results and the discussion of these experiments are presented in Sections~\ref{section:results} and~\ref{sec:discussion}. Finally, conclusions and future work are discussed in Section~\ref{section:conclusions}.

\section{SEIRS model}
\label{section:seir}
\subsection{Classical SEIRS model}
\begin{figure*}[t!]
	\centering
	\includegraphics[width=0.85\linewidth]{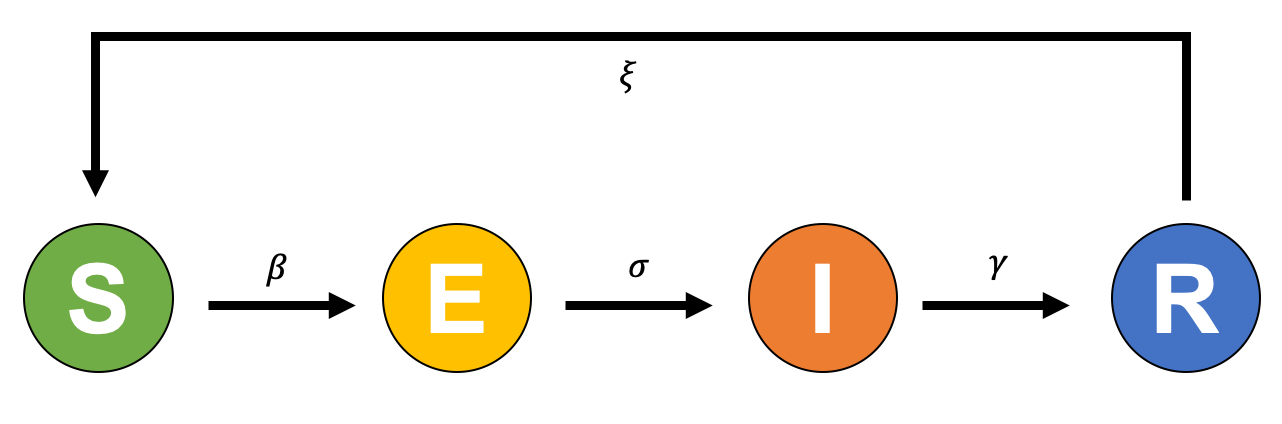}
	\caption{Key variables and parameters in the SEIRS model representing the compartments Susceptible (S), Exposed (E), Infectious (I), and Recovered (R). Modified from \citet{idm:2020}.}
	\label{fig:seirmodel}
\end{figure*} 


The SEIRS equations that govern virus infection dynamics categorise the population into four compartments:  Susceptible, Exposed, Infectious or Recovered. See Figure~\ref{fig:seirmodel} for an illustration of the rates that control how a person moves between these compartments. %
The infection rate, $\beta$, controls the rate of spread which represents the probability of transmitting disease between a susceptible and an exposed individual (someone who has been infected but is not yet infectious). The incubation rate, $\sigma$, is the rate of exposed individuals becoming infectious (the average duration of incubation is $1/\sigma$). The recovery rate, $\gamma = 1/{T_D}$, is determined by the average duration, $T_D$, of infection. For the SEIRS model, $\xi$ is the rate at which recovered individuals return to the susceptible state due to loss of immunity.




Vital dynamics can be added to a SEIRS model, by including birth and death rates represented by $\mu$ and $\nu$, respectively. To maintain a constant population, one can make the assumption that $\mu = \nu$, however, in the general case, the system of ordinary differential equations can be written: 
\begin{subequations} \label{SEIRS-eqn}
\begin{eqnarray}
\frac{\partial S}{\partial t} & = &\mu N  - \frac{\beta SI}{N} + \xi R - \nu S, \\
\frac{\partial E}{\partial t} & = &\frac{\beta SI}{N} - \sigma E - \nu E, \\ 
\frac{\partial I}{\partial t} & = &\sigma E - \gamma I - \nu I, \\
\frac{\partial R}{\partial t} & = &\gamma I  - \xi R- \nu R
\end{eqnarray}
\end{subequations}
where $S(t)$, $E(t)$, $I(t)$ and $R(t)$ represent the number of individuals in the susceptible, exposed (infected but not yet infectious), infectious and recovered compartments respectively. At time~$t$, the total number of individuals in the population under consideration is given by $N(t)=S(t)+E(t)+I(t)+R(t)$. If the birth and death rates are the same, $N$ remains constant over time.

\subsection{\gls{Extended SEIRS} model}
In this study, the SEIRS model is extended in two ways. First, we introduce diffusion terms to govern how people move throughout the domain, thereby incorporating spatial variation into the model. Second, we associate a group with each person, indicated by the index $h\in\{1, 2, \dots, \mathcal{H}\}$. This indicates the person has gone to work or school, gone shopping, gone to a park or stayed at home, for example, and transmission rates for each group can be set according to the risk of being in offices, schools, shopping centres, outside, or at home. 
These modifications to the SEIRS equations result in the following system of equations:
\begin{subequations}
\label{compart-SEIRS-diff-comb-eqn} 
\begin{align} 
\frac{\partial S_h}{\partial t} & = \mu_h N_h  - \frac{S_h \sum_{h'}(\beta_{h\, h'}   I_{h'}) }{ N_{h}} 
+  \xi_h R_h - \nu_h^S S_h - \sum_{h'=1}^{\cal H} \lambda_{hh'}^S S_{h'} 
+ \nabla \cdot (k_h^S \nabla S_h)   , 
\label{compart-SEIRS-diff-comb-eqn1} \\[3mm]
 \frac{\partial E_h}{\partial t}  & =  \frac{S_h \sum_{h'}(\beta_{h\, h'}   I_{h'}) }{ N_{h}}
-  \sigma E_h - \nu_h^E E_h  
- \sum_{h'=1}^{\cal H} \lambda_{hh'}^E E_{h'}  + \nabla \cdot (k_h^E \nabla E_h) ,
\label{compart-SEIRS-diff-comb-eqn2}
\\[2mm]
\frac{\partial I_h}{\partial t}  & =   \sigma E_h - \gamma_h I_h - \nu_h^I I_h 
- \sum_{h'=1}^{\cal H} \lambda_{hh'}^I I_{h'} 
+ \nabla \cdot (k_h^I \nabla I_h) ,
\label{compart-SEIRS-diff-comb-eqn3} 
\\[2mm]
\frac{\partial R_h}{\partial t}  & =   \gamma_h I_h  - \xi_h R_h 
- \nu_h^R R_h
- \sum_{h'=1}^{\cal H} \lambda_{hh'}^R R_{h'}  
+ \nabla \cdot (k_h^R \nabla R_h) , 
\label{compart-SEIRS-diff-comb-eqn4} 
\end{align}
\end{subequations} 
in which the subscript $h$ represents which group an individual is associated with. Instead of having scalar values for each compartment, we now have fields: $S_h(\bm{\omega},t)$, $E_h(\bm{\omega},t)$, $I_h(\bm{\omega},t)$ and $R_h(\bm{\omega},t)$, where the people associated with group~$h$ for the susceptible, exposed, infectious and recovered compartments, respectively, vary in space, $\bm{\omega}$, and time,~$t$. 
The transmission terms $\beta_{hh'}$ govern how the disease is transmitted from people in groups~$h'\in\{1,2,\dots,\mathcal{H}\}$ to people in group~$h$. The terms involving $\lambda_{hh'}^{(\cdot)}$ are interaction terms that control how people move between the groups describing the various locations/activities for the compartment given in the superscript. These values could, for example, control whether people in the school group move into the home group. When moving from one group to another, the individual remains in the same compartment. Describing the spatial variation, the diffusion coefficients for each compartment are given by $k_h^{(\cdot)}$. The birth rate for a group is $\mu_h$ and the death rate is set for each compartment and group, where, for example, $\nu_h^{S}$ is the death rate of group~$h$ for the susceptible compartment. The term $\sigma$ represents the rate at which some of the people in the 
exposed compartment, $E$, transfer to the infectious compartment, $I$.  
The recovery rate is now:
\begin{equation}\label{gammas-eqn}
\gamma_h = \frac{1}{T_{D_h}}, 
\end{equation}
in which $T_{D_h}$ are the average durations of infections in infection groups $I_h$. Therefore the infectious rates become:
\begin{equation}
\beta_{h h} = \gamma_h R0_{h}, \quad h\in \{1, 2, \ldots \mathcal{H}\}. 
\end{equation}
Here we assume $\beta_{h h'} =0$  when $h \neq h'$. This assumption means that a person in the Home group cannot infect someone in the Mobile group (as the former will be at home and the latter will be outside of the home), and, likewise, a person in the Mobile group cannot infect someone in the Home group. 

An eigenvalue problem can be formed by placing an eigenvalue, $\lambda_0$, in front of the terms $\sigma E_h$ in equations~\eqref{compart-SEIRS-diff-comb-eqn2} and~\eqref{compart-SEIRS-diff-comb-eqn3}, and by setting all four time derivatives to zero in equations \eqref{compart-SEIRS-diff-comb-eqn}. In addition, this term will need to be linearised. To model the beginning of the virus outbreak, a possible way of linearising is shown here:
\begin{equation}
    \label{diffuse-eig-approx}  
    \frac{S_h^g \sum_{h'}(\beta_{h\, h'}   I_{h'}) }{ N_{h}}
    \approx  
    \sum_{h'}(\beta_{h\, h'}  I_{h'}), \quad
    \forall h\in\{ 1,2,\ldots,\mathcal{H} \}. 
\end{equation} 
The eigenvalue is equivalent to the reciprocal of $R0$, 
that is $R0 = \frac{1}{\lambda_0}$. 

We remark that the system of equations~\eqref{compart-SEIRS-diff-comb-eqn} is similar to the neutron transport equations and comment that codes written to solve nuclear engineering problems could be reapplied to virus modelling without much modification.

\subsection{Extended SEIRS model for two groups}
\label{subsec:extendedSEIRS}
\begin{figure*}[htbp]
\centering
	\includegraphics[trim = {2cm 1cm 1cm 1cm},clip,width=0.75\linewidth]{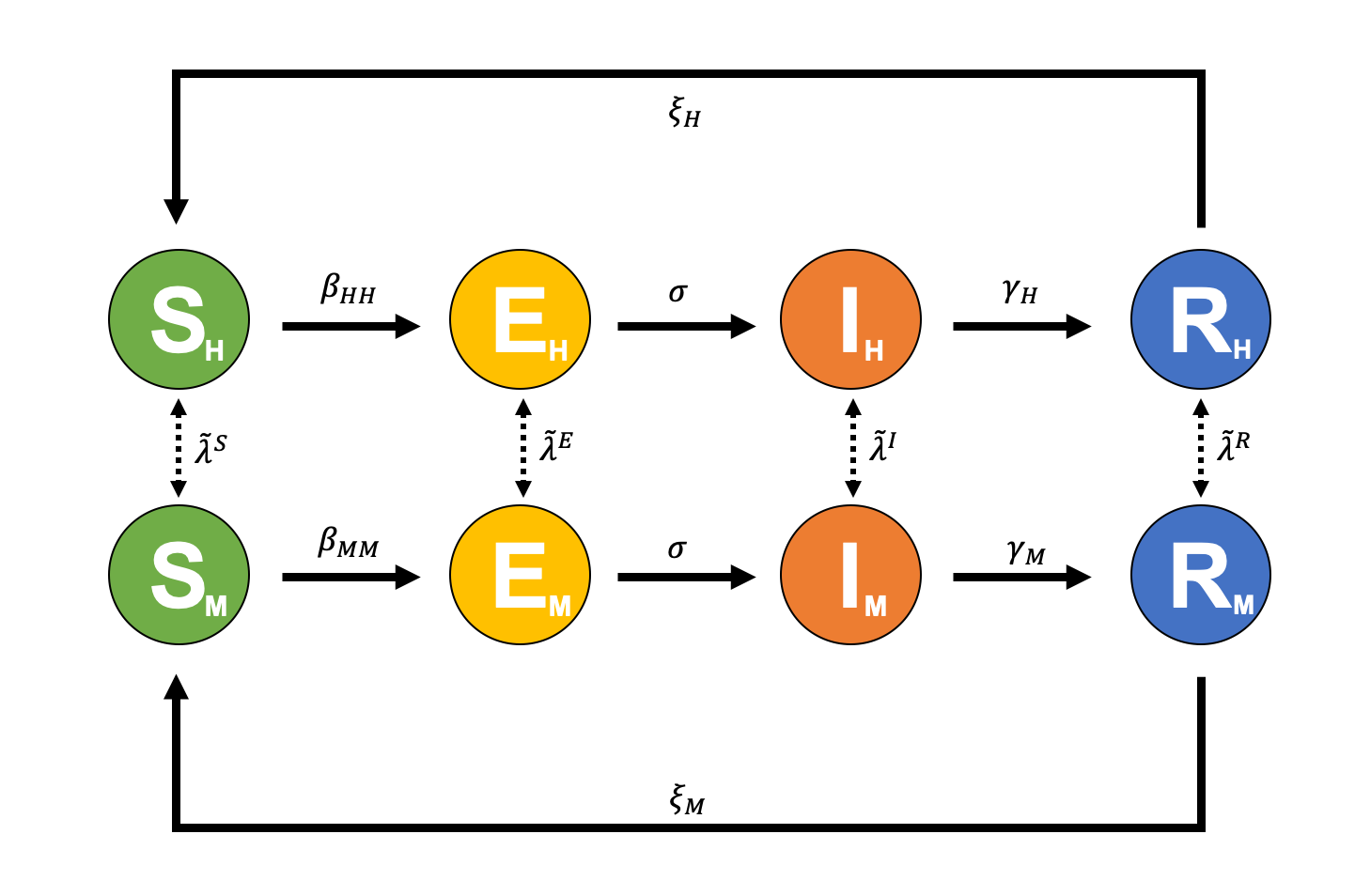}
\caption{Movement between compartments Susceptible (S), Exposed (E), Infectious (I) and Recovered (R), and groups Home (H) and Mobile (M) for the extended SEIRS model. The spatial variation is not represented here, just movement between compartments and groups. The movement between home and mobile groups is defined by $\tilde{\lambda}^{(.)}$.}
\label{extended_SEIRS_diagram}
\end{figure*}

As said in the introduction, the area of reduced-order modelling is new to virus modelling, so we choose a simple test case to try to understand the application of these methods to virus modelling. In this paper, we restrict ourselves to the specific case where there are two possible and distinct groups in addition to the SEIRS compartments. The groups comprise people who remain at home (`Home', $H$), and others who are mobile and can move to riskier surroundings (`Mobile', $M$). The index representing the group, $h$, has therefore two values: $h\in\{H,M\}$. For this case, the transmission terms between Home and Mobile must be zero, so $\beta_{HM}=0$ and $\beta_{MH}=0$. This is because an individual at Home cannot infect someone in the Mobile group and vice versa as they will not be near one another. We wish interaction terms $\lambda_{hh'}^{(\cdot)}$, which control how people move from Home to Mobile groups and vice versa, to be such that conservation is obeyed. In other words, the number of people leaving the Home group (for a given compartment) must equal the number of people entering the Mobile group (for that compartment). On inspection of equation~\eqref{compart-SEIRS-diff-comb-eqn1}, for group $h=H$, we can see that people moving between the Home and Mobile groups in the susceptible compartment will be $-\lambda_{HH}^S S_H-\lambda_{HM}^S S_M$. From equation~\eqref{compart-SEIRS-diff-comb-eqn1}, for group $h=M$, people moving between the Home and Mobile groups in the susceptible compartment is given by the terms $-\lambda_{MM}^S S_M-\lambda_{MH}^S S_H$. To enforce that the number of people leaving $S_H$ is equal to the number of people joining $S_M$, the interaction coefficients can be set as follows:
\begin{equation}\label{eq:interactions}
    \lambda_{HH}^S = -\lambda_{MH}^S\,, \quad \lambda_{MM}^S = -\lambda_{HM}^S \quad \text{and} \quad \lambda_{HH}^S = \lambda_{MM}^S\ .
\end{equation}
Suppose $\lambda_{HH}^S =: \tilde{\lambda}^S$, then we can say that the number of people leaving~$S_M$ (joining if $\tilde{\lambda^S}<0$) is $\tilde{\lambda}^S(S_H-S_M)$ and the number of people joining~$S_H$ (leaving if $\tilde{\lambda^S}<0$) is $\tilde{\lambda^S}(S_H-S_M)$. Similar relationships hold for the other three compartments, i.e.~replace the superscript~$S$ in equations~\eqref{eq:interactions} with $E$, $I$ and $R$ in turn. See Figure~\ref{extended_SEIRS_diagram} for an illustration of how people move between compartments and groups in this extended SEIRS model. \citeauthor{Radulescu2020}~\cite{Radulescu2020} uses a similar approach to model a small college-town which has seven locations (medical centre, shops, university campus, schools, parks, bars and churches) all with appropriate transmission rates.

The spatial variation is discretised on a regular grid of $N_X \times N_Y \times N_Z$ control volume cells. 
The point equations can be recovered by choosing $N_X = N_Y = N_Z = 1$. We use a 5~point stencil and second-order differencing of the diffusion operator, as well as backward Euler time stepping. We iterate within a time step, using Picard iteration, until convergence of all nonlinear terms and evaluate these nonlinear terms at the future time level. To solve the linear system of equations we use Forward Backward Gauss-Seidel (FBGS) for each variable in turn, and once convergence has been achieved, Block FBGS is used to obtain overall convergence of the eight linear solutions. This simple solver is sufficient to solve the relatively small problems presented here. 

The parameters $\beta_{hh'}$, $\sigma$, $\gamma_{h}$ and $\xi_{h}$, were chosen based on parameters observed in the UK, similar to~\citet{nadler2020epidemiological} who also estimated the parameters from data, this time for the SIR equations. According to the \citet{govuk2020}, the incubation period is between 1 and 14 days, with a median of 5 days. Here, an incubation rate of 4.5 days is used, which is within the range of observed COVID-19 incubation periods in the UK. The SEIRS model presented here is flexible, however, meaning that it could be applied to other regions with different parameters. More details about the configuration of the SEIRS model and the results obtained are given in Section \ref{sec:domain}.

\section{Methods}
\label{section:Methods}

\subsection{Bidirectional Long Short-term Memory networks}
\label{subsec:mBiLSTM}
The LSTM network comprises three gates: input ($\mathbf{i}_{t_{k}}$), forget ($\mathbf{f}_{t_{k}}$), and output ($\mathbf{o}_{t_{k}}$); a block input, a single cell $\mathbf{c}_{t_{k}}$, and an output activation function. This network is recurrently connected back to the input and the three gates. Due to the gated structured and the forget state, the LSTM is an effective and scalable model that can deal with long-term dependencies \citep{hochreiter1997long}. The vector equations for a LSTM layer are:

\begin{equation}\label{eq:lstm}
\begin{aligned}
{\bf i}_{t_{k}}\ &= \phi({\bf W}_{xi}\mathbf{x}_{t_{k}} + {\bf W}_{Hi}{\bf H}_{t_{k-1}} + {\bf b}_{i})\\
{\bf f}_{t_{k}}\ &= \phi({\bf W}_{xf}\mathbf{x}_{t_{k}} + {\bf W}_{Hf} {\bf H}_{t_{k-1}} + {\bf b}_{f})\\
{\bf o}_{t_{k}}\ &= \phi({\bf W}_{xo}\mathbf{x}_{t_{k}} + {\bf W}_{Ho}{\bf H}_{t_{k-1}} + {\bf b}_{o})\\
{\bf c}_{t_{k}}\ &= {\bf f}_{t_{k}} \circ {\bf c}_{t_{k-1}} + {\bf i}_{t_{k}} \circ \tanh({\bf W}_{xc}\mathbf{x}_{t_{k}} + {\bf W}_{Hc}{\bf H}_{t_{k-1}} + {\bf b}_{c})\\
{\bf H}_{t_{k}}\ &= {\bf o}_{t_{k}}\circ \tanh({\bf c}_{t_{k}})\\
\end{aligned}
\end{equation}
where $\phi$ is the sigmoid function, $\mathbf{W}$ are the weights, $\mathbf{b}_{i,f,o,c}$ are the biases for the input, forget, output gate and the cell, respectively, $\mathbf{x}_{t_{k}}$ is the layer input, $\mathbf{H}_{t_{k}}$ is the layer output and $\circ$ denotes the entry-wise multiplication of two vectors.

The idea of BDLSTMs comes from bidirectional RNN \citep{schuster1997bidirectional}, in which sequences of data are processed in both forward and backward directions with two separate hidden layers. BDLSTMs connect the two hidden layers to the same output layer. %
The forward layer output sequence is iteratively calculated using inputs in a forward sequence, $\overrightarrow{\mathbf{H}_{t_{k}}}$, from time $t_{k-n}$ to $t_{k-1}$, and the backward layer output sequence, $\overleftarrow{\mathbf{H}_{t_{k}}}$, is calculated using the reversed inputs from $t_{k-1}$ to $t_{k-n}$. The layer outputs of both sequences are calculated by using the equations in \eqref{eq:lstm}. The BDLSTM layer generates an output vector $\mathbf{u}_{t_{k}}$:

\begin{equation}
\mathbf{u}_{t_{k}} = \psi(\overrightarrow{\mathbf{H}_{t_{k}}},\overleftarrow{\mathbf{H}_{t_{k}}})
\end{equation}
where $\psi$ is a concatenating function that combines the two output sequences. 

\subsubsection{Prediction with BDLSTM}
\begin{figure*}[t]
	\centering
	\includegraphics[width=0.9\linewidth]{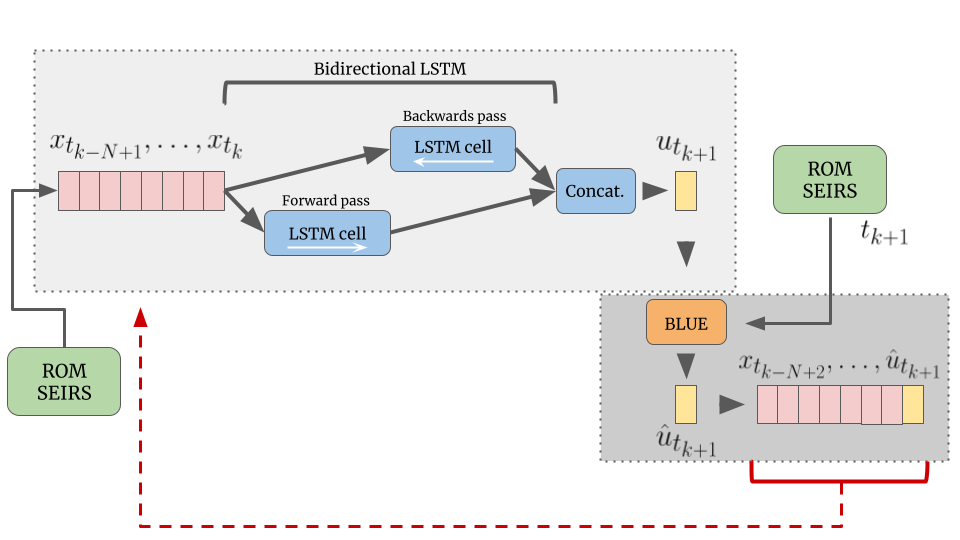}
	\caption{Predictive LSTM \gls{$f^{BDLSTM+BLUE}$} for a sequence of two time levels. Top-left: off-line bidirectional LSTM network. Bottom-right: data-correction of the prediction. The Best Linear Unbiased Estimation (BLUE) is used to data-correct the prediction of the network. One time level corresponds to 10 time-steps of the original SEIRS solution.}
	\label{fig:PredBDLSTM}
\end{figure*}

The prediction workflow with the BDLSTM is presented in Figure~\ref{fig:PredBDLSTM}. While LSTMs are known for producing time-series predictions, the workflow introduces a data-corrected step. This step improves the accuracy of those predictions. The BDLSTM network \gls{$f^{BDLSTM}$} is a function trained off-line to predict $t_{k+1}$ given the previous $N$ time-levels from the latent vector $\mathbf{x}$, that represents the ROM:

\begin{equation}
f^{BDLSTM}:\mathbf{x}_{t_{k-N+1}},\dots, \mathbf{x}_{t_{k}} \to  \mathbf{\tilde{u}}_{t_{k+1}}.
\end{equation}

Once the network is able to predict the solution $\tilde{\mathbf{u}}_{t_{k+1}}$, this is joined to the solutions at $u_{t_{k-N}}, u_{t_{k-N+1}}, \ldots, u_{t_{k}}$, to create $\mathbf{u}_{p}$. The prediction vector $\mathbf{u}_{p}$ is then optimised online using the Best Linear Unbiased Estimator (\gls{BLUE}):

\begin{equation}
    \hat{\mathbf{u}}_{p} = \bar{\mathbf{u}}_{p} + \mathbf{C}_{\mathbf{u}_{p}\mathbf{v}}\mathbf{C}^{-1}(\mathbf{v}-\bar{\mathbf{v}})
\end{equation}
where $\hat{\mathbf{u}_{p}}$ is the data-corrected prediction, $\bar{\mathbf{u}_{p}}$ is the mean of the vector $\mathbf{u}_{p}$ over time, $\mathbf{v}$ and $\bar{\mathbf{v}}$ are the observations and mean of the observations over time, respectively, $\mathbf{C}_{\mathbf{u}_{p}\mathbf{v}}$ is the covariance between $\mathbf{u}_{p}$ and observations $\mathbf{v}$, and $\mathbf{C}$ is the covariance of the observations. 
The first entry of $\mathbf{u}_{p}$ is dropped and the new vector is used to make a prediction of $t_{k+2}$. This is an iterative process. Thus, the data-corrected BDLSTM is defined by:

\begin{equation}
    f^{BDLSTM+BLUE}:\mathbf{x}_{t_{k-N+1}},\dots, \mathbf{x}_{t_{k}} \to  \mathbf{\hat{u}}_{p}
\end{equation}

In the prediction with the BDLSTM workflow, before performing a PCA on the original dataset, we normalised the values of each compartment by their corresponding means and standard deviation. This step was not done for the predictive GAN.

\subsection{Generative adversarial network }
\label{subsec:mPredGAN}

Proposed by \citet{goodfellow:14}, Generative Adversarial Networks (GANs), are unsupervised learning algorithms capable of learning dense representations of the input data and are intended to be used as a generative model, i.e.~they are capable of learning the distribution underlying the training dataset and able to generate new samples from this distribution. The training of the GAN is based on a game theory scenario in which the generator network $G$ must compete against an adversary. The generator network $G$ directly produces time-sequences from a random distribution as input (latent vector $\mathbf{z}$):
\begin{equation}
    G:\mathbf{z}\sim\mathcal{N}(0,I_L) \to \mathbf{y}_{GAN} \in \mathbb{R}^{N\times M}
\end{equation}
where $\mathbf{y}_{GAN}$ is an array of $N$ time sequences with $M$ dimensions, $L$ is the size of the latent vector, and $I_L$ is an identity matrix of size $L$. The discriminator network $D$ attempts to distinguish between samples drawn from the training data, the ROM, and samples drawn from the generator, considered as fake. The output of the discriminator $D(\mathbf{y})$ represents the probability that a sample came from the data rather than a “fake” sample from the generator, and the vector $\mathbf{y}$ represents ``real'' samples of the principal components from the ROM. The output of the generator $G(\mathbf{z})$ is a sample from the distribution learned in the dataset. Equations \eqref{eq:Ld} and \eqref{eq:Lg} show the loss function of the discriminator and generator, respectively:

\begin{equation}
    \label{eq:Ld}
   L_D = -\mathbb{E}_{y \sim p_{data}(y)}[\log(D(\mathbf{y}))] - \mathbb{E}_{z \sim p_{z}(z)}[\log(1- D(G(\mathbf{z})))]
\end{equation}

\begin{equation}
    \label{eq:Lg}
   L_G = \mathbb{E}_{z \sim p_{z}(z)}[\log(1- D(G(\mathbf{z})))]
\end{equation}

\subsubsection{Predictions with GAN}
\label{subsubsec:predgan}
\begin{figure*}[t]
	\centering
	\includegraphics[width=0.9\linewidth]{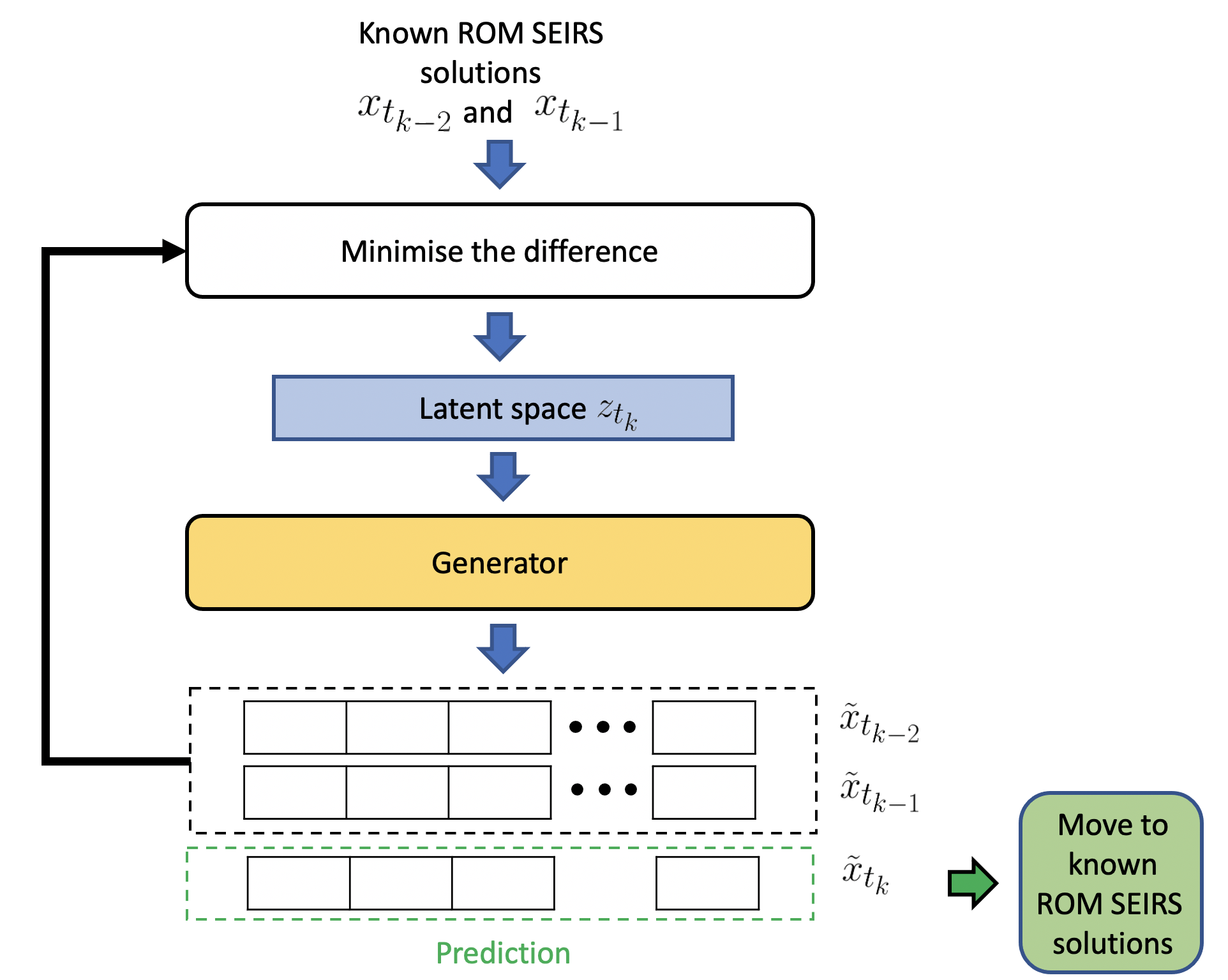}
	\caption{Workflow of \gls{$f^{PredictiveGAN}$} for a sequence of three time levels ($N=3$). Adapted from \citet{Silva2020}.}
	\label{fig:PredGANc}
\end{figure*}

To make predictions in time using a GAN, an algorithm named Predictive GAN~\cite{Silva2020} is used. The network is trained to generate data at a sequence of $N$ time levels from $t_{k-N+1}, \dots, t_{k}$ no matter at which point in time $k$ is. In other words, the network will generate data that represents the dynamics of $N$ consecutive time levels. Following that, given known solutions from time levels $t_{k-N+1}$ to $t_{k-1}$, the input of the generator $\mathbf{z}$ can be optimised to produce solutions at time levels $t_{k-N+1}$ to $t_{k}$. Hence the new prediction is the solution at time $t_{k}$. To predict the next time level, having known solutions at $t_{k-N+2}$ to $t_{k-1}$ and the newly predicted solution at time $t_{k}$, we can predict the solution at time level $t_{k+1}$. The process repeats until predictions have been obtained for all the desired time levels.  Figure~\ref{fig:PredGANc} illustrates how the predictive GAN works.

In this work, for a GAN that has been trained to predict $N$ time levels, $G(\mathbf{z}_{t_{k}})$ is defined as     
\begin{equation}\label{xgan}
G(\mathbf{z}_{t_{k}}) = \left[ \begin{array}{c} 
(\mathbf{\tilde{x}}_{t_{k-N+1}})^T \\[1mm]
(\mathbf{\tilde{x}}_{t_{k-N+2}})^T \\[1mm]
\vdots \\[1mm]
(\mathbf{\tilde{x}}_{t_{k}})^T \\[1mm]
\end{array}
\right],
\end{equation}
where $(\mathbf{\tilde{x}}_{t_k})^T= [\tilde{x}^1_{t_k}, \tilde{x}^2_{t_k}, \cdots , \tilde{x}^{M}_{t_k}]$ and it represents a predicted low dimension space of the extended SEIRS model states at time level $t_k$.  $M$ is the number of principal components used in the ROM, and $\tilde{x}^i_{t_k}$ represents the predicted $i$th principal component at time level $t_k$.   

Considering we have solutions at time levels from $t_{k-N+1}$ to $t_{k-1}$, denoted by $\{\mathbf{x}_{t_j}\}_{j=k-N+1}^{k-1}$, then to predict the solution at time level $t_k$ we perform an optimisation defined as:  
\begin{equation}\label{eq:opt}
    \begin{gathered}
    \mathbf{z}_{t_k}  = \argmin_{\mathbf{z_{t_k}}} \mathcal{L}(\mathbf{z}_{t_k}),  \\
    \mathcal{L}(\mathbf{z}_{t_k})  =  \sum_{j=k-N+1}^{k-1}\, \left(\mathbf{x}_{t_j} - \mathbf{\tilde{x}}_{t_j} \right)^T \bm{W}_\alpha \left(\mathbf{x}_{t_j} - \tilde{\mathbf{x}}_{t_j} \right),  
    \end{gathered} 
\end{equation}
where $\bm{W}_\alpha$ is a square matrix of size $M$ whose diagonal values are equal to the principal components weights, all other entries being zero. It is worth noticing that only the time levels from $t_{k-N+1}$ to $t_{k-1}$ are taken into account in the functional which controls the optimisation of $\mathbf{z}_{t_k}$. The new predict time level $t_{k}$ is added to the known solutions $\mathbf{x}_{t_k}$ = $\tilde{\mathbf{x}}_{t_k}$, and the converged latent variables $\mathbf{z}_{t_k}$ are used to initialise the latent variables at the next optimisation to predict time level $t_{k+1}$. The process repeats until all time levels are predicted. It is worth mentioning that the gradient of Equation \eqref{eq:opt} can be calculated by automatic differentiation \citep{wengert:64, linnainmaa:76, baydin:17}. In other words, the error generated by the loss function is backpropagated in Equation \eqref{eq:opt} through the generator. 

Finally, the predictive GAN function is defined by:
\begin{equation}
f^{PredictiveGAN}: \mathbf{x}_{t_{k-N+1}},\dots, \mathbf{x}_{t_{k-1}} \to \mathbf{\tilde{x}}_{t_{k-N+1}},\dots, \mathbf{\tilde{x}}_{t_{k}}
\end{equation}
and, $f^{BDLSTM+BLUE}$ and $f^{PredictiveGAN}$ are both iterative processes that represent the forecast functions from the BDLSTM+BLUE method and the Predictive GAN method, respectively.

\subsection{Feed-forward network}

To show that $f^{BDLSTM}$ and $f^{PredictiveGAN}$ are better suited for temporal sequences, a deep feed-forward network (FFN) is employed. Feed-forward networks are neural networks where the connection between nodes do not form a cycle and the network moves only forward \citep{goodfellow2016deep}, unlike recurrent neural networks. Here, we trained a simple deep FFN and further improve its performance with BLUE, named $f^{FFN+BLUE}$:

\begin{equation}
    f^{FFN+BLUE}:\mathbf{x}_{t_{k-N+1}},\dots, \mathbf{x}_{t_{k}} \to  \mathbf{\hat{u}}_{p}
\end{equation}

\section{Results}
\label{section:results}

The following section presents the test case, the parameters used in the extended SEIRS model (shown in Section \ref{subsec:extendedSEIRS}), and the predictions of the two digital twin models of the spread of the COVID-19 infection for this idealised scenario explained in Section \ref{subsec:mBiLSTM} and \ref{subsec:mPredGAN}, respectively. The models are general, however, and could be applied to mode complex scenarios. The first digital twin is based on a bidirectional LSTM and the second is based on a predictive GAN model. Both systems were implemented using TensorFlow \citep{tensorflow:2015} and the Keras wrapper \citep{chollet2015keras} in Python. 
\subsection{Test case}
\label{sec:domain}

The domain of the test case occupies an area measuring 100km by 100km and is subdivided into 25 regions as shown in Figure~\ref{fig:regions}. Those labelled as~1 are regions into which people do not travel and the region labelled as~2 is where homes are located. People in the home group remain at home in region~2, and people in the mobile group can travel anywhere in regions labelled~2 or~3.  Within this domain, the modified SEIRS equations will model the movement of people around the domain as well as determining which compartment and group the people are in at any given time.  People can be in one of four compartments: Susceptible, Exposed, Infectious or Recovered, and for each of these, people can either be at Home or Mobile. To model the spatial variation, diffusion is used as the transport process.

\begin{figure*}[ht]
	\centering
    \begin{tikzpicture}

\filldraw[fill=gray!20!white, draw=green!40!black] (2,0) rectangle (3,5);
\filldraw[fill=gray!20!white, draw=green!40!black] (0,2) rectangle (2,3);
\filldraw[fill=gray!20!white, draw=gray] (3,2) rectangle (5,3);

\draw[step=1cm,color=gray, very thick] (0,0) grid (5,5);

\node at (0.5,+4.5) {1};
\node at (1.5,+4.5) {1};
\node at (2.5,+4.5) {3};
\node at (3.5,+4.5) {1};
\node at (4.5,+4.5) {1};

\node at (0.5,+3.5) {1};
\node at (1.5,+3.5) {1};
\node at (2.5,+3.5) {3};
\node at (3.5,+3.5) {1};
\node at (4.5,+3.5) {1};

\node at (0.5,+2.5) {3};
\node at (1.5,+2.5) {3};
\node at (2.5,+2.5) {3};
\node at (3.5,+2.5) {3};
\node at (4.5,+2.5) {3};

\node at (0.5,+1.5) {1};
\node at (1.5,+1.5) {1};
\node at (2.5,+1.5) {3};
\node at (3.5,+1.5) {1};
\node at (4.5,+1.5) {1};

\node at (0.5,+0.5) {1};
\node at (1.5,+0.5) {1};
\node at (2.5,+0.5) {2};
\node at (3.5,+0.5) {1};
\node at (4.5,+0.5) {1};

\fill [red] (2.75, +0.25) circle(4pt);
\end{tikzpicture} 
	\caption{Cross-shaped area in a domain of $100 \text{km}\times100 \text{km}$. The grey regions represent where people can travel. The red dot indicates a location at which comparison will be made between the two experiments using BDLSTM and GAN.}
	\label{fig:regions}
\end{figure*}

Now we must set the coefficients for the extended SEIRS model. For both transient simulations and steady state eigenvalue equations, for regions~2 and~3, the diffusion coefficients  are set to:
\begin{equation}
k_h^c=\frac{2.5L^2}{T_\text{one day}}, \qquad  k_h^c= 0.05\frac{2.5L^2}{T_\text{one day}} \quad \forall h\in\{H,M\},\  \forall c\in\{S,E,I,R\}
\end{equation} 
respectively, in which $L$ is a typical length scale. Here, $L$ is taken as the length of the domain, i.e.~100km. For region~1, all diffusion coefficients are zero, thus no people will move into this region, see Figure~\ref{fig:regions}.
$R0_{h}, \; h\in\{H,M\}$ are the 
the average number of people in group $h$ a person within group $h$ infects while in that group. 
In this example,  $R0_{H}=0.2$ for people at Home ($h=H$), and $R0_{M}=10$ for Mobile people ($h=M$). 
If one solves an eigenvalue problem, using these values of $R0_h$, starting from 
an initial uninfected population, then the 
resulting overall $R0$ is $R0=7.27$. That is 
one person at the infectious stage of the virus can infects on average $7.27$ other people. 
The death rate is assumed to equal the birth rate, given by: 
\begin{equation}
\mu =\frac{1}{(60 \times 365\times T_\text{one day})} = \nu, 
\label{fortan-1}
\end{equation}
where the average age at death is taken to be $60$ years and $T_\text{one day}$ is the number of seconds in one day. The rate at which recovered individuals return to the susceptible state due to loss of immunity for both Home and Mobile groups is defined as: 
\begin{equation}
\xi_h =  \frac{1}{(365 \times T_\text{one day})}.
\end{equation}

The interaction terms or intergroup transfer terms, $\lambda^{(\cdot)}_{hh'}$, govern how people in a particular compartment move from the home to the mobile group, or vice versa. The aim is that most people will move from home to mobile group in the morning, travel to locations in regions~2 or~3 and return home later on in the day. To achieve this, the values $\lambda^{(\cdot)}_{hh'}$ depend on other parameters, as now described. Night and day is defined through the variable:
\begin{equation}
            R_{DAY}= 0.5 \sin \left( \frac{2\pi t}{T_\text{one day}}\right)  + 0.5\,, 
\label{fortan-2}
\end{equation}
in which $t$ is time into the simulation. For region 2 (see Figure~\ref{fig:regions}):
\begin{equation}
N_{H\,aim}=1000 (1-R_{DAY}) + 1000, \quad 
{N_M}_{aim}=0, \quad 
\Lambda_{H,H} = \frac{1000}{T_\text{one day}}\,. 
\label{fortan-3}
\end{equation}
$N_{H\,aim}$  and ${N_M}_{aim}$ 
can be thought of as the total number of people 
that we aim to have in the $H$ and $M$ groups in region~2 (i.e.~where there are homes). This results in a
pressure to move people from  their homes during the day and back into them during the night time when they return home. Thus, $\Lambda_{H,H}$ is set in such a way as to move people out of their homes on time scale of $\frac{1}{1000}$ of a day. For all other regions:
\begin{equation}
N_{H\,aim}=0, \quad 
{N_M}_{aim}=0, \quad 
\Lambda_{H,H} = 0. 
\label{fortan-4}
\end{equation} 
For time dependent problems, a forcing term 
is defined as:  
\begin{eqnarray}
{\cal S}_{H2M}& =&0.5+0.5 \; \sign(F), 
\label{fortan-7}
\end{eqnarray} 
where 
\begin{eqnarray}
F&=&\frac{ N_H - N_{H\,aim} } {\max\{ \epsilon, N_H , N_{H\,aim} \} }, \label{fortan-6} 
\end{eqnarray} 
in which $\sign(F)=1$ if $F\geqslant0$, otherwise $\sign(F)=-1$.
With this definition of ${\cal S}_{H2M}$, 
in equation \eqref{fortan-7}, for time-dependent problems, we can define the intergroup transfer terms as follows:

\begin{align}
\lambda_{H,H}^S &=  0.01\Lambda_{H,H}   \; {\cal S}_{H2M} F;  &\lambda_{H,H}^E=\lambda_{H,H}^I=\lambda_{H,H}^R=\lambda_{H,H}^S,
\label{fortan-8} \\
\lambda_{M,M}^S &=  -\Lambda_{H,H}    (1-{\cal S}_{H2M}) F;  &\lambda_{M,M}^E=\lambda_{M,M}^I=\lambda_{M,M}^R=\lambda_{M,M}^S,
\label{fortan-9}\\
\lambda_{H,M}^S &=  \Lambda_{H,H}  (1-{\cal S}_{H2M}) F ;  &\lambda_{H,M}^E=\lambda_{H,M}^I=\lambda_{H,M}^R=\lambda_{H,M}^S,
\\
\lambda_{M,H}^S &=  -0.01\Lambda_{H,H}   {\cal S}_{H2M} \; F ; &\lambda_{M,H}^E=\lambda_{M,H}^I=\lambda_{M,H}^R=\lambda_{M,H}^S.
\label{fortan-10}
\end{align}
For eigenvalue problems,  the parameters are defined as follows:
\begin{eqnarray}
r_{ratio}&=&25.65; \label{fortan-11}\\ 
r_{switch}& =& \left\{ \begin{array}{l} 1\quad \text{in region 1}\\0 \quad\text{elsewhere.}\end{array} \right.\nonumber
\end{eqnarray}
The parameter $r_{switch}$ switches on the home location in the equations below:
\begin{equation}
 \Lambda_{H,H} = \frac{r_{switch}}{T_\text{one day}}; \;\;\;\; \Lambda_{M,M} = 10000 \frac{(1-r_{switch})}{T_\text{one day}}.
\label{fortan-12}
\end{equation}
The intergroup transfer coefficients are set to be
\begin{align}
\lambda_{H,H}^S &=  \frac{1}{\epsilon};\quad    \lambda_{H,H}^R=\lambda_{M,M}^S=\lambda_{M,M}^R=\lambda_{H,H}^S,
\label{fortan-13}\\
\lambda_{H,H}^E &= \lambda_{H,H}^I =\Lambda_{H,H}    + \Lambda_{M,M}, 
\label{fortan-14}\\
\lambda_{M,M}^E &= \lambda_{M,M}^I =\Lambda_{H,H}  r_{ratio}. 
\label{fortan-15}\\
\lambda_{H,M}^S & = \lambda_{H,M}^E =\lambda_{H,M}^I =\lambda_{H,M}^R =-\Lambda_{H,H}  r_{ratio}, 
\label{fortan-16}\\
\lambda_{M,H}^S &= \lambda_{M,H}^E =\lambda_{M,H}^I =\lambda_{M,H}^R =-\Lambda_{H,H}.
\label{fortan-17}
\end{align}  
This defines all the parameters required for the extended SEIRS model.

We are thus modelling the daily cycle of night and day for the transient calculations, in which there is a pressure for mobile people to go to their homes at night, and there will be many people leaving their homes during the day moving to the mobile group. For region~2, the average ratio of the number of people at home to the number of people that are mobile from the transient calculations during the first 10~days of the simulation is used to form the ratio $r_{ratio}$. This ratio is then used in the steady-state eigenvalue calculations to enforce consistency with the transient calculations. However, acknowledging the difference in the steady-state and time-dependent diffusion terms we scale the former by a factor of $0.05$ as shown in Equation~\eqref{fortan-11} above. The coefficient $\frac{1}{\epsilon}$, where $\epsilon= 10^{-10}$, was added onto the diagonal of all the $S$ and $R$ equations (as shown above) to effectively set their values to approximately zero as they play no role in the eigenvalue calculations. This enables only minor modifications to be made to the transient code, to give the eigenvalue problem. 


The domain of the numerical simulation is divided into a regular mesh of $10\times10$ cells. As there are four compartments and two groups in this problem, there will be eight variables for each cell in the mesh per time step, which gives a total number of $800$ variables per time step. The total time of the transient simulation is $3888\times 10^3$ seconds, or $45.75$ days, with a time step of $\Delta t = 1000$ seconds resulting in $3880$ time levels. Each control volume is assumed to have 2000 people in the home region cells and all other fields are set to zero, so only susceptible people are non-zero at home initially. This is with the exception that we assume that $0.1\%$ of people at 
home has been exposed to the virus and will thus develop an infection. 

\begin{figure*}[t!]
	\centering
	\includegraphics[width=0.9\linewidth]{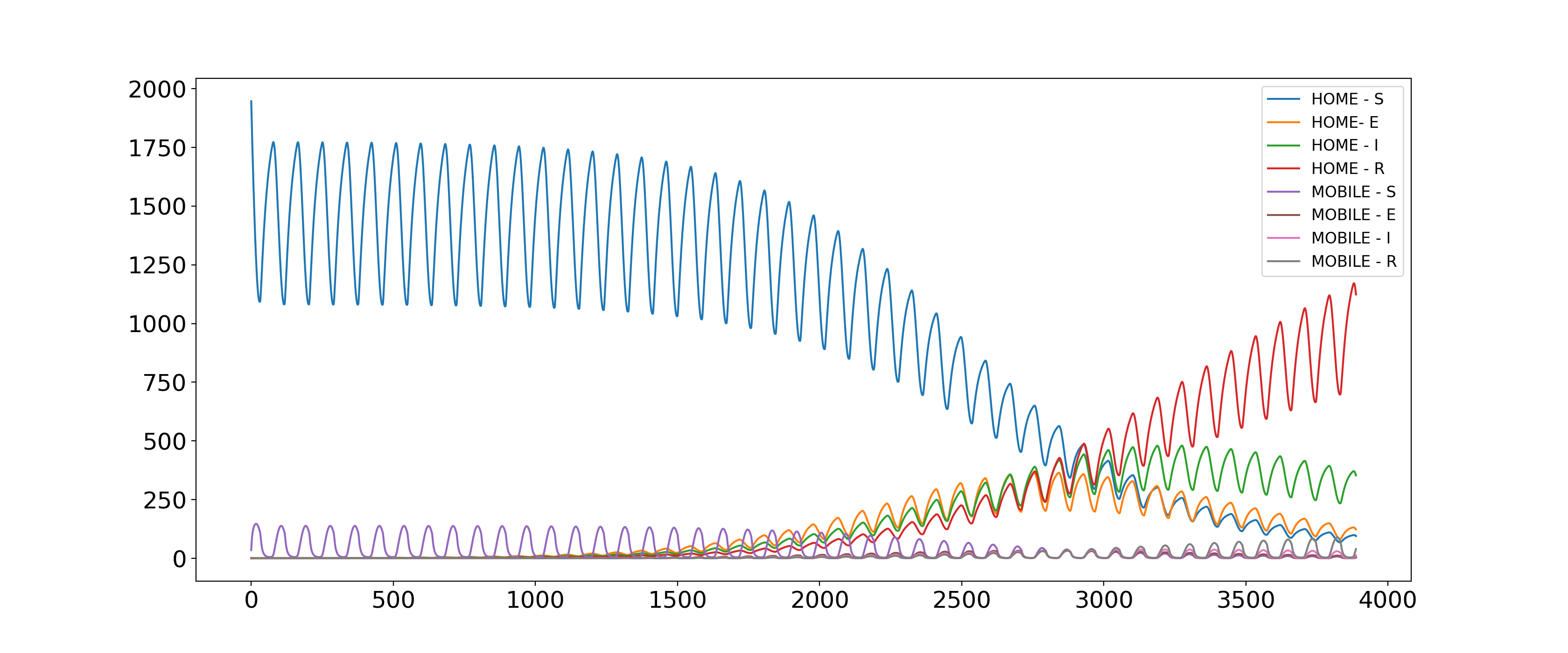}
	\caption{Total number of people in each compartment and group versus time.}
	\label{fig:SEIR_time_series}
\end{figure*}

The $S$, $E$, $I$, $R$ fields for people at home and mobile are shown in Figure~\ref{fig:SEIR_time_series} for the default transient configuration over 45 days. The daily cycle might, for instance, start at about 6 am (e.g.~$t=0$), say, where people start to leave their homes. People have started to leave their homes, become mobile and start to diffuse through the domain. This continues towards the end of the day where they have moved further away from their homes. However, at midnight they make their way back to their homes and thus, with a relatively small spread of the virus near the homes. Notice that at this time level, a small percentage of the population is exposed, infectious or recovered, and the rest is susceptible to $S$. We see the daily cycle of people moving from their homes to becoming mobile and we also see the gradual increase in the number of people in the exposed, infectious and recovered compartments for both mobile and home groups. Notice that the number of exposed and infectious people increases rapidly in this simulation and then starts to decrease because the number of susceptible people decreases. That is, recovered people gradually increases and they are immune.

\begin{figure*}[t!]
	\centering
	\includegraphics[width=0.9\linewidth]{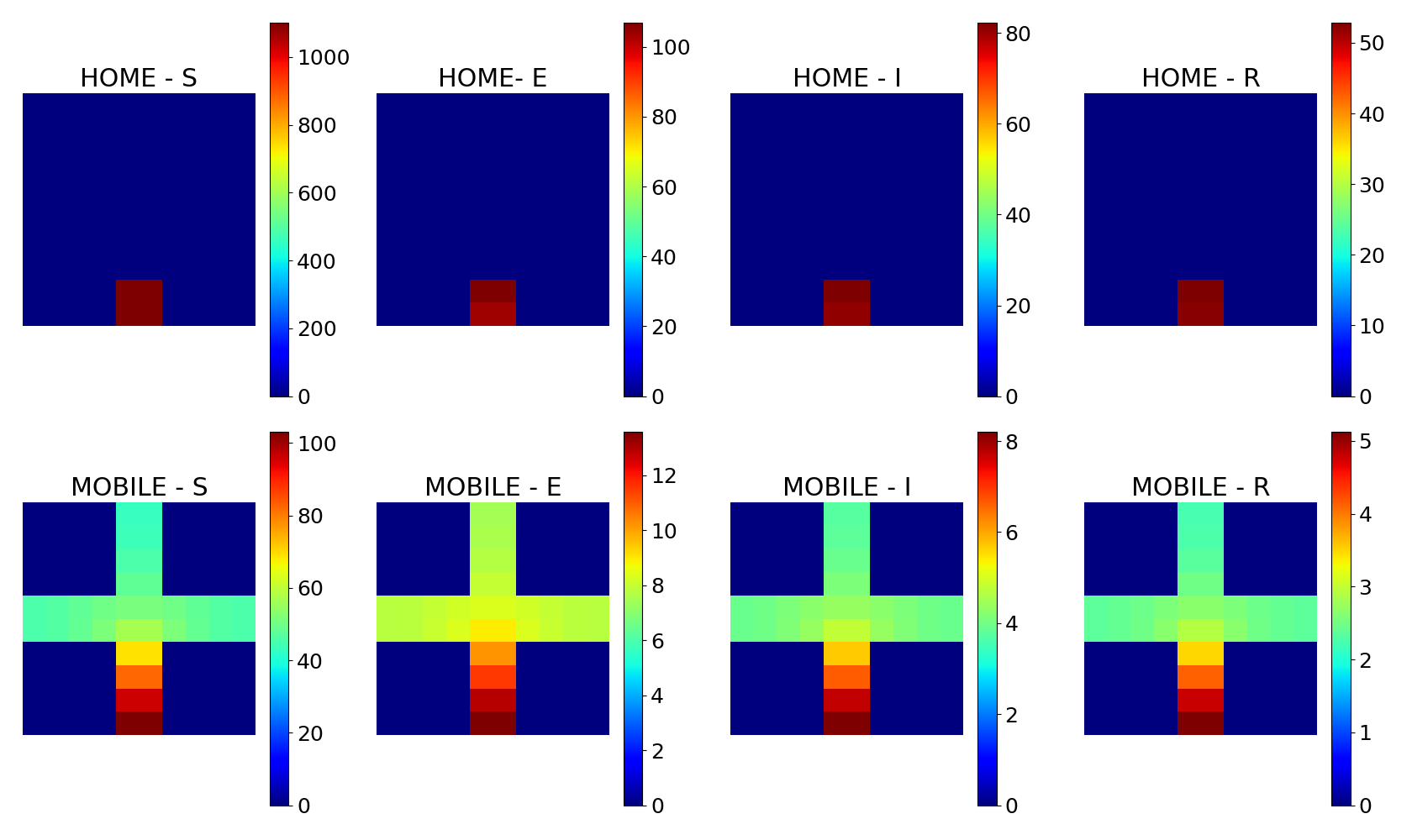}
	\caption{Spatial variation of the test case domain after $2\times10^6$ seconds for the Home (top) and Mobile groups (bottom) and the S, E, I and R compartments (left to right).}
	\label{fig:2000TS}
\end{figure*}

\subsection{Reduced order modelling}

A principal component analysis (PCA) is performed on the $800$ variables ($100$ points in space in $4$ compartments and $2$ groups), to obtain a low-dimensional space in which the predictive GAN and BDLSTM operate. The first $15$ principal components were chosen, as they represent $>99.9\%$ of the variance. Both methods sample data every 10 time-steps from the PCs. Thus, both methods have access to $388$ time levels. The time lag in both experiments is 8, as this configuration roughly represents a cycle (one day) of the original extended SEIRS simulation.
The main goal of both methods is to be able to act as surrogate models for the extended SEIRS model, producing predictions in a much faster time than is required to solve the extended SEIRS model itself (assuming the latter is sufficiently demanding).

\begin{figure*}[htb]
	\centering
	\includegraphics[width=0.7\linewidth]{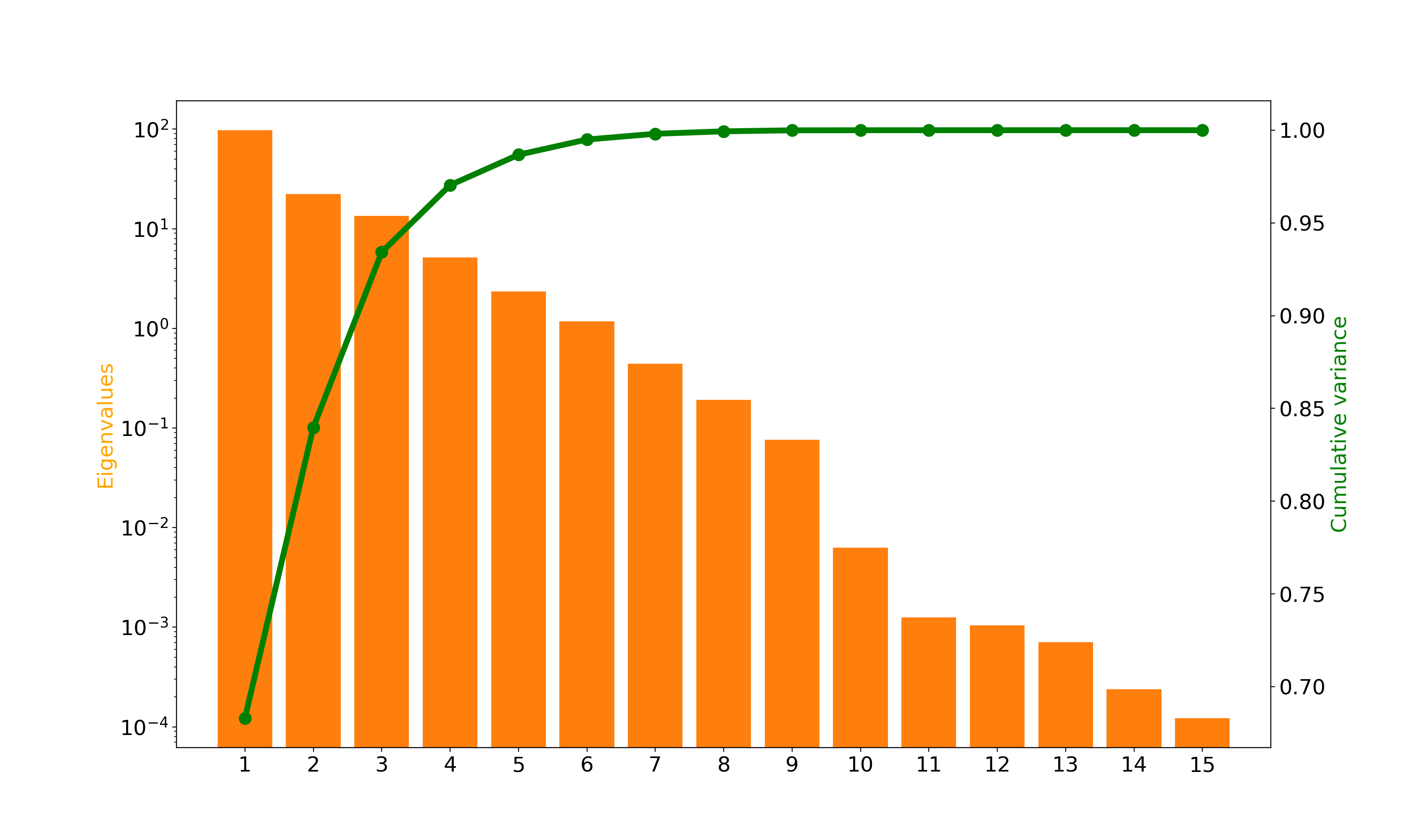}
	\caption{Eigenvalues (left) and normalised cumulative sum of the variance (right) of the first 15 components.}
	\label{fig:PCAsv}
\end{figure*}

\subsection{Bidirectional Long short-term memory network}
\label{subsec:predlstm}

\begin{figure*}[ht]
	\centering
	\includegraphics[width=1.0\linewidth]{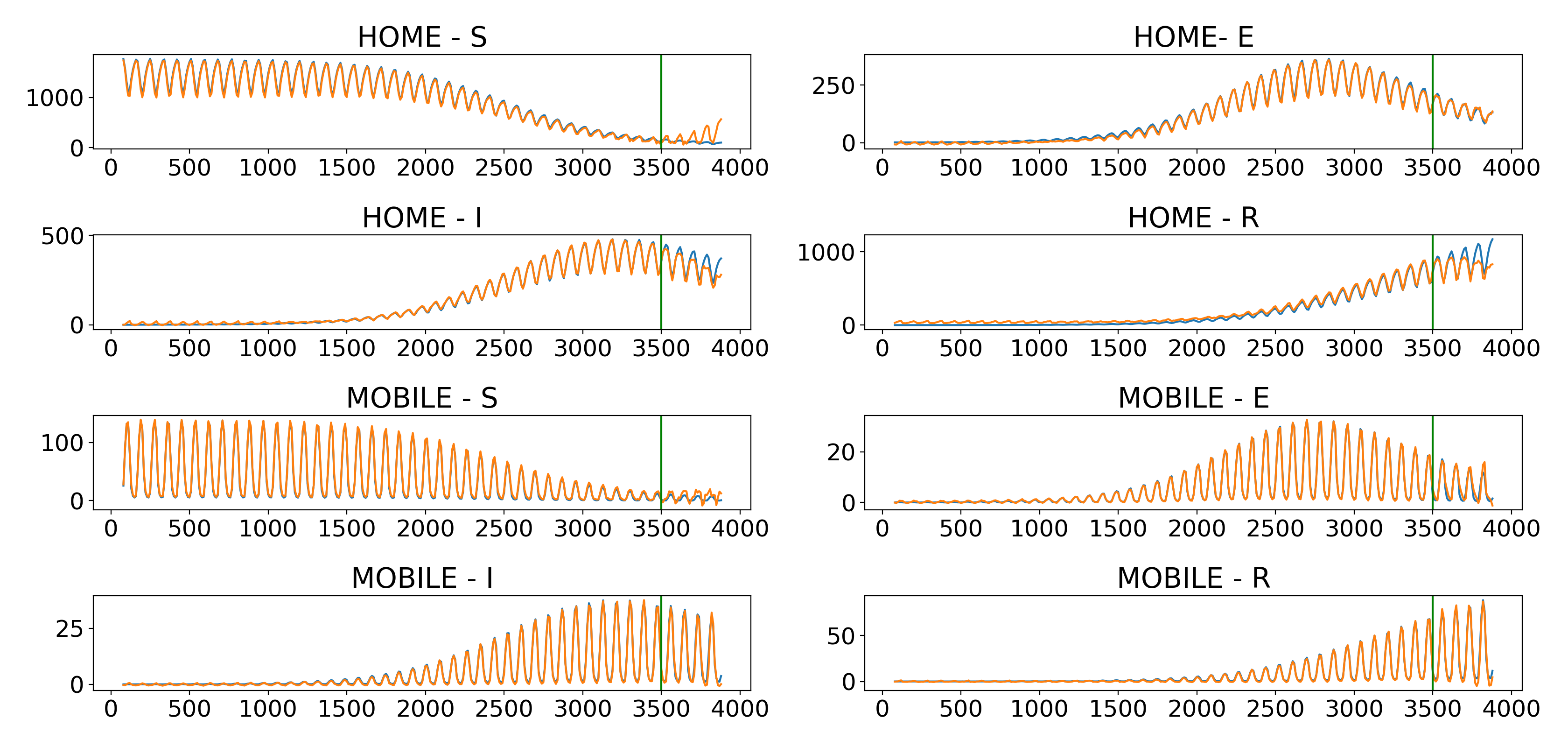}
	\caption{The $f^{BDLSTM}$ prediction (orange) over time of the outcomes of the infection (in number of people) in one point (marked as a red circle in Fig \ref{fig:regions}) of the mesh starting at time step $0$. The predictions are off-line, not data-corrected and have a sliding window of 8 time-steps and use the data from the original dataset (blue) to predict the next one. The green line shows the start of the test data.}
	\label{fig:lstm1by1}
\end{figure*}

The network $f^{BDLSTM}$ is trained using the previous 8 time levels $t_{k-7}, t_{k-6}, \ldots, t_{k}$ (namely 80 time-steps of the original extended SEIRS simulation) to generate the next one $t_{k+1}$ (10 time steps ahead of the original extended SEIRS simulation), with a time interval of 10 time steps. The network is trained using 90\% of the available data, reserving the remaining 10\% for testing. Figure~\ref{fig:lstm1by1} depicts the prediction of one time-step, at a single point of the domain, using data from the original simulation, once $f^{BDLSTM}$ is trained. This is a validation that the model can make accurate predictions on both the training data and the test data.

The BDLSTM architecture is based on \citet{cui2018deep} and $f^{BDLSTM}$ was trained for 500 epochs using a grid search of hyperparameters including hidden nodes in the LSTM layer, batch sizes, and dropouts.

\begin{figure*}[ht]
	\centering
	\includegraphics[width=1.0\linewidth]{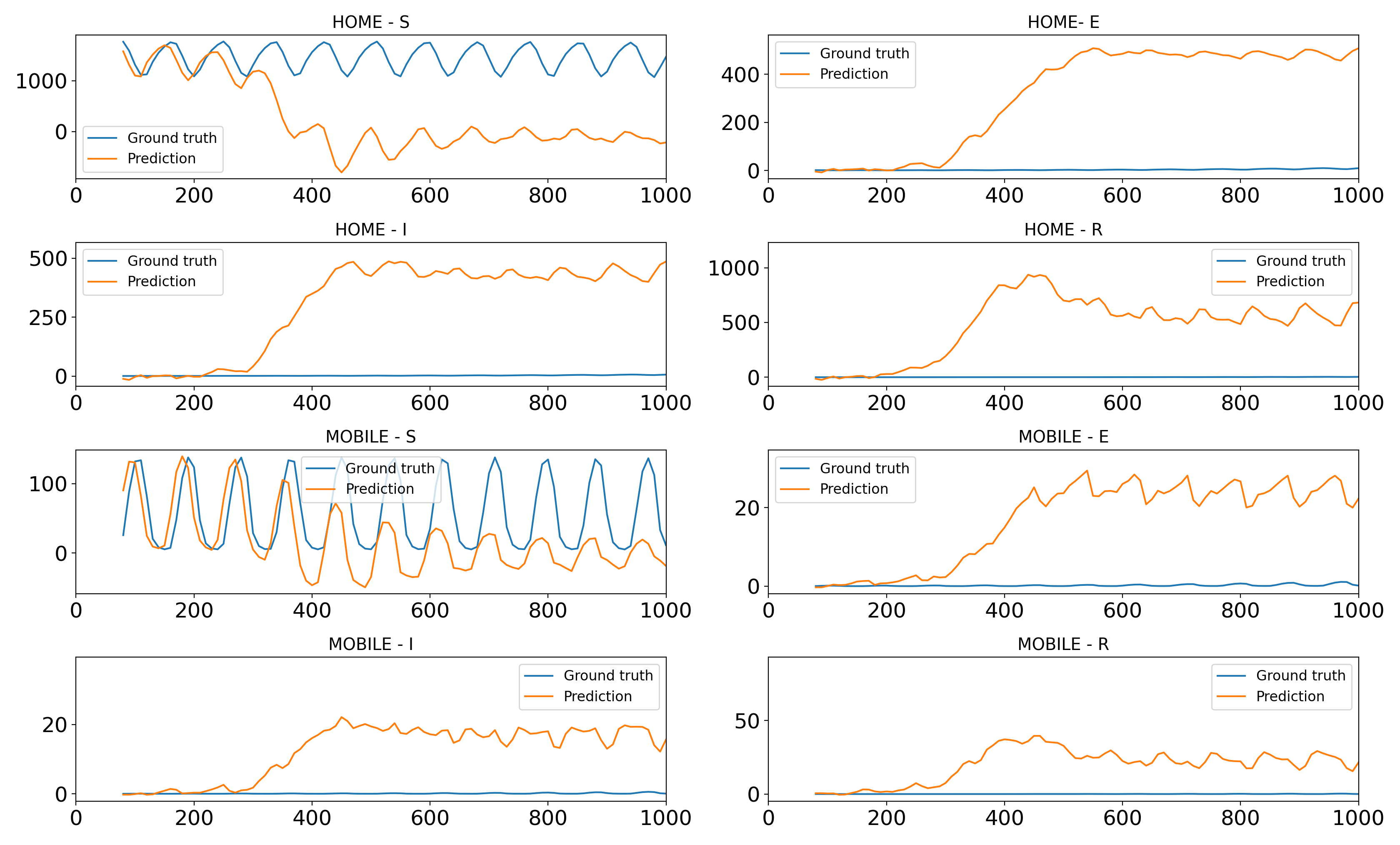}
	\caption{The $f^{BDLSTM}$ prediction, over time, of the outcomes of the infection (in number of people) in one point (marked as a red circle in Fig \ref{fig:regions}) without any data-correction from time-step $0$. The predictions from $f^{BDLSTM}$ act iteratively like an input for the prediction of the following time-step.}
	\label{fig:lstm_diverging}
\end{figure*}

Without including data-correction (Figure~\ref{fig:lstm_diverging}), the predictions from $f^{BDLSTM}$ start after diverging $\sim$ 30 iterations. This means that $f^{BDLSTM}$ does not diverge greatly from the original dataset before $\sim$ 30 cycles of input-output, without external information. Therefore, the prediction by $f^{BDLSTM}$ needs to be data-corrected to align with the dynamics of the extended SEIRS solution.

\begin{figure*}[t!]
\centering
\begin{subfigure}[t]{1\textwidth}
    \centering
	\includegraphics[width=0.8\linewidth]{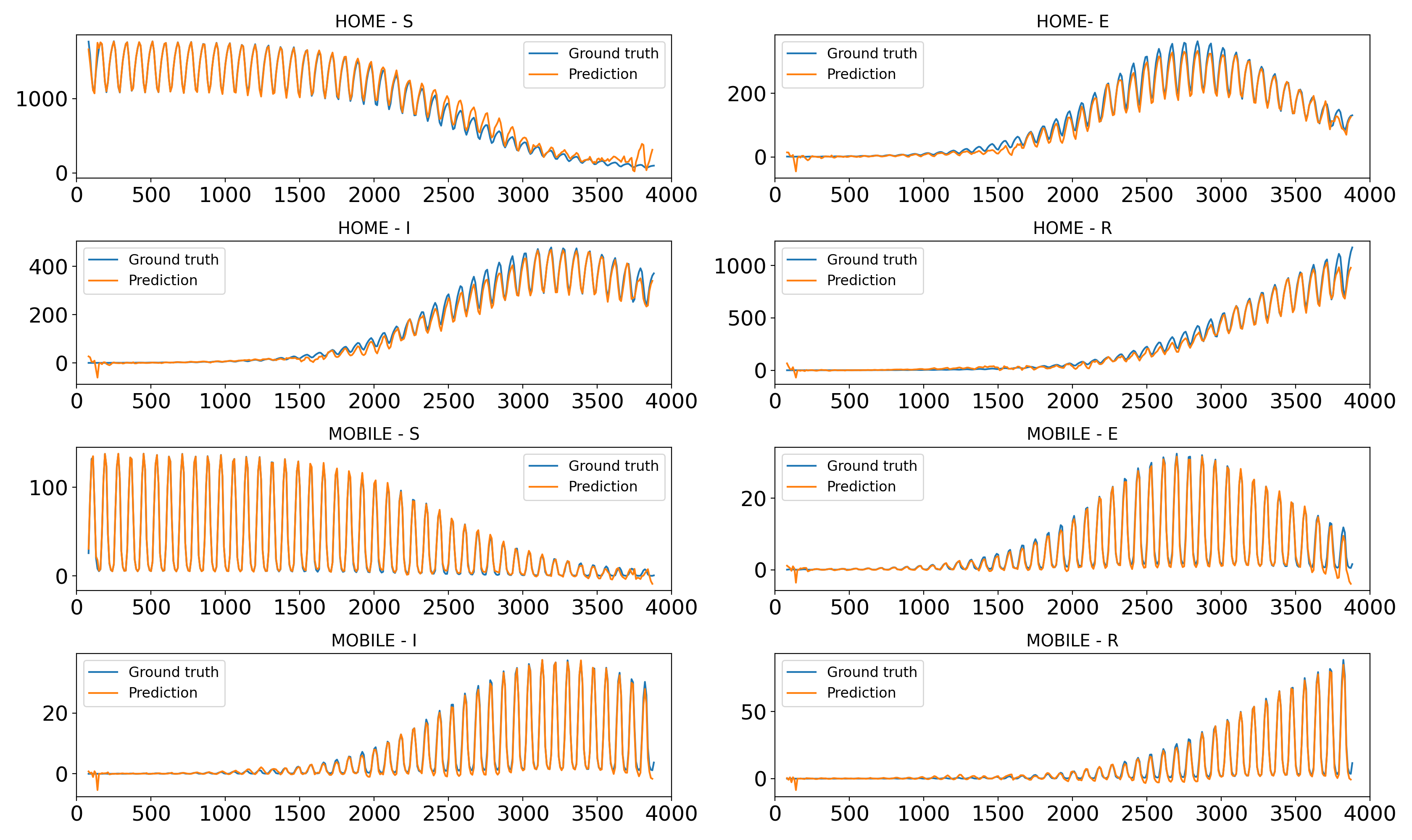}\label{fig:bdlstm0-end}
	\caption{Starting at $9\times10^4$ seconds.}
\end{subfigure}
\begin{subfigure}[t]{1\textwidth}
    \centering
	\includegraphics[width=0.8\linewidth]{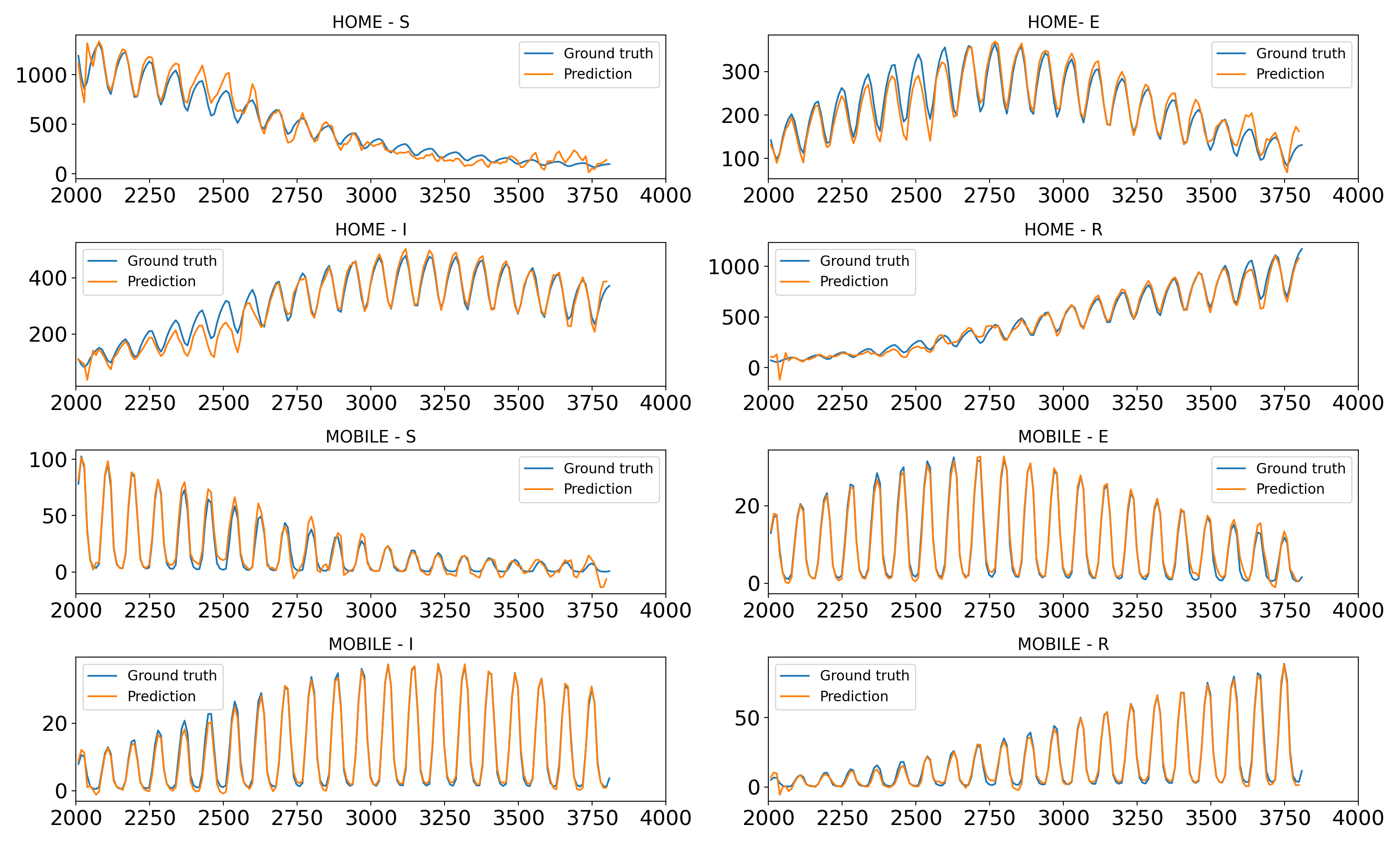}
	\label{fig:bdlstm2000-end}
	\caption{Starting at $2\times10^6$ seconds.}
\end{subfigure}
\caption{$f^{BDLSTM+BLUE}$ prediction (in number of people) at one point (marked as a red circle in Fig \ref{fig:regions}) of the domain over time starting from different time levels.}
\label{fig:bdlstmPreds}
\end{figure*}

The data-corrected prediction by the BDLSTM, $f^{BDLSTM+BLUE}$ starting from time step $90$ ($9\times 10^4$ seconds), is shown in Figure~\ref{fig:bdlstmPreds}a. Each cycle in the curves corresponds roughly to a period of one day. Figure~\ref{fig:bdlstmPreds}b depicts the data-corrected prediction every 10 time-steps starting from time-step 2000 of the simulation ($2 \times 10^6$ seconds). Comparable results are obtained at other points of the mesh. In both cases, $f^{BDLSTM+BLUE}$ struggles at predicting the Susceptible compartments in both Home and Mobile groups. The $f^{BDLSTM+BLUE}$ performs poorly at predicting the initial values in both cases starting from the beginning of the dataset and from $t=2000$ ($2 \times 10^6$ seconds).

\subsection{Prediction using GAN}
\label{subsec:rprgan}

A predictive GAN, $f^{PredictiveGAN}$, is applied to the spatial variation of COVID-19 infection, to make predictions based on training using data from the numerical simulation. The generator and discriminator are trained using a sequence of 9 time levels with a time interval of 10 time steps between them. The first 8 time levels are used in the optimisation process, described in Section \ref{subsubsec:predgan}, and the last time level is used in the prediction. The network is trained using all time steps of the numerical simulation.

The GAN architecture is based on DCGAN \citep{radford:15}. The generator and discriminator are trained for $55,000$ epochs. The 9 time levels are given to the networks as a two-dimensional array with nine rows and fifteen columns. Each row represents a time level and each column is a principal component from PCA (a low dimensional representation of the simulation states). Although it is not an image, it can be represented as one. The DCGAN can take advantage of the time dependency of the two-dimensional array (the image), as the simulation states for the first time level is in the first row, for the second time level is in the second row, and so forth. During the optimisation process in each iteration of $f^{PredictiveGAN}$, the singular values from the SVD are used as weights in the Equation \eqref{eq:opt}.

The prediction in $f^{PredictiveGAN}$ is performed by starting with 8 time levels from the numerical simulation and using the generator to predict the ninth. During the next iteration, the last prediction is used in the optimisation process and this is repeated until the end of the simulation. It is worth mentioning that after 8 iterations the $f^{PredictiveGAN}$ works only with data from the predictions. Data from the numerical simulation is used only for the starting points.

\begin{figure*}[t!]
\centering
\begin{subfigure}[t]{1\textwidth}
    \centering
	\includegraphics[width=0.8\linewidth]{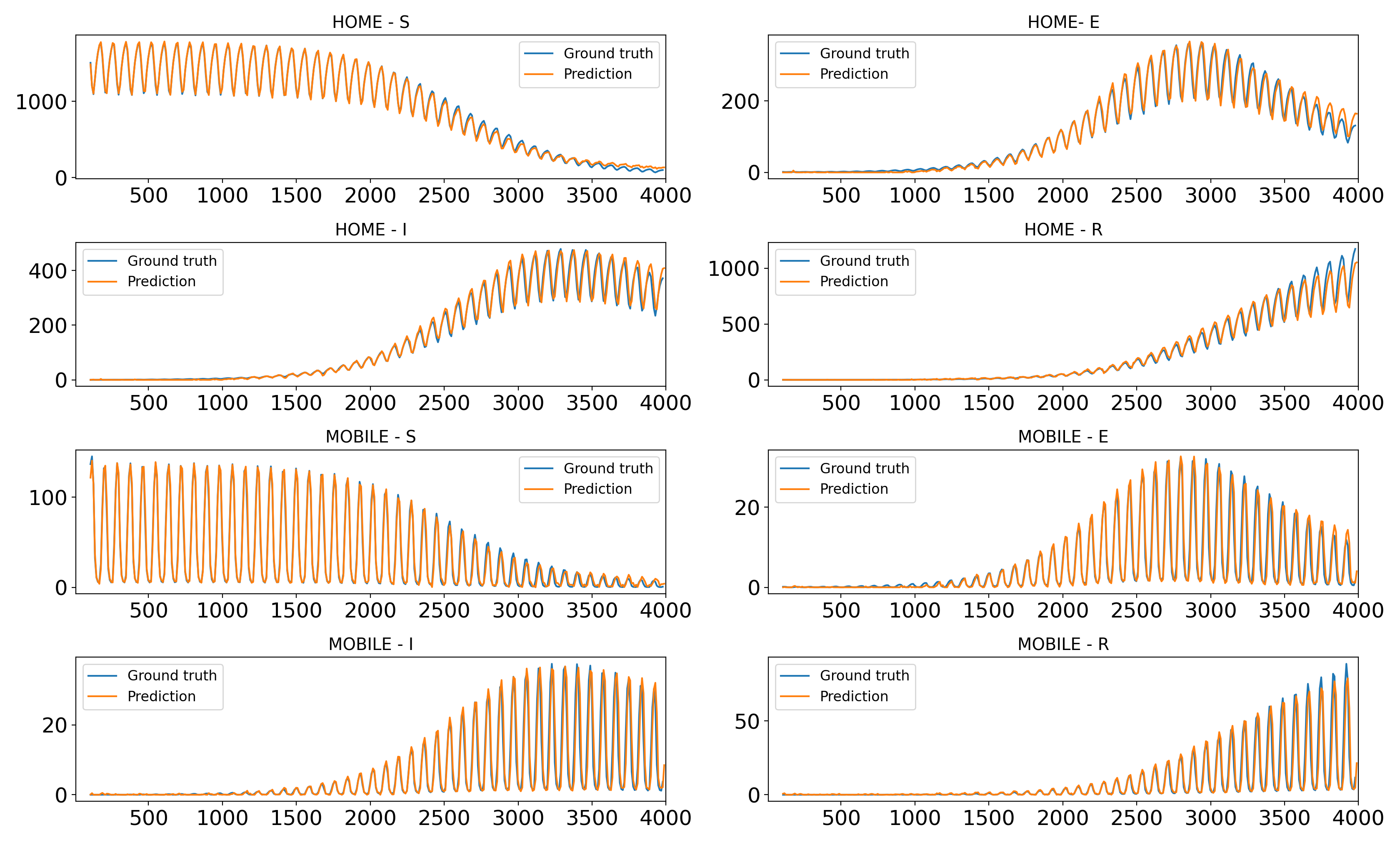}
	\caption{Starting at $9\times10^4$ seconds.}
	\label{fig:prgan0-end}
\end{subfigure}
\begin{subfigure}[t]{1\textwidth}
    \centering
	\includegraphics[width=0.8\linewidth]{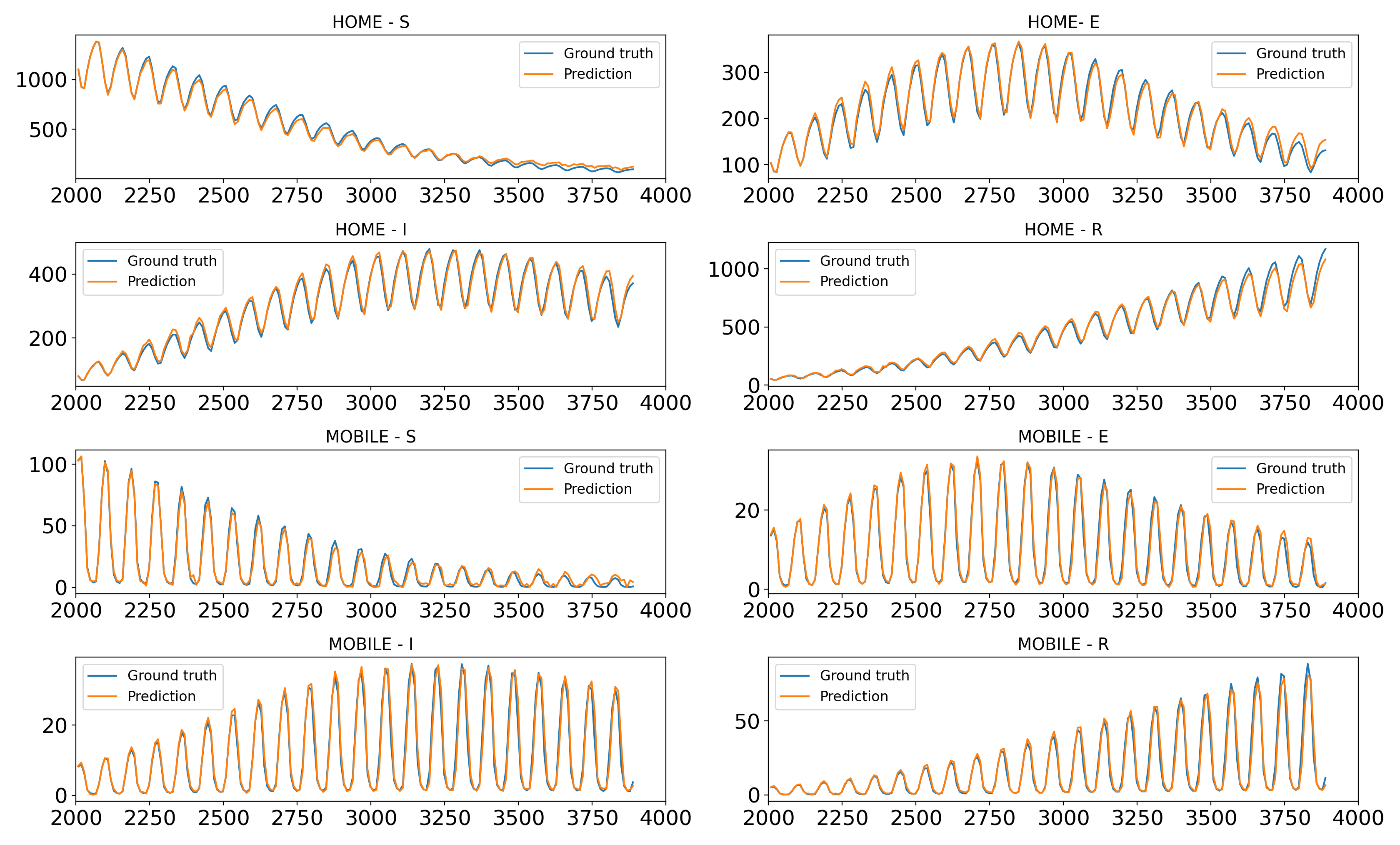}
	\caption{Starting at $2\times10^6$ seconds.}
	\label{fig:prgan2000-end}
\end{subfigure}
\caption{$f^{PredictiveGAN}$ prediction (in number of people) at one point (marked as a red circle in Fig \ref{fig:regions}) of the domain over time starting from different time levels.} 
\label{fig:prganPreds}
\end{figure*}

Figure \ref{fig:prganPreds}a shows the prediction over time of $f^{PredictiveGAN}$ but in one point of the mesh (bottom-right corner of region 2 shown in Figure~\ref{fig:regions}). Each cycle in the curves corresponds to a period of one day. The process is repeated this time with the simulation starting at time step $2\times10^3$ ($2\times10^6$ seconds). The result over time for one point of the mesh (bottom-right corner of region 2) is presented in Figure~\ref{fig:prganPreds}b. Comparable results regarding the error in the prediction are obtained at other points of the mesh, therefore we do not present them here. We can notice from Figure~\ref{fig:prganPreds} that $f^{PredictiveGAN}$ can reasonably predict the outcomes of the numerical model.       

\subsection{Feed forward network}

\begin{figure*}[t!]
\centering
\begin{subfigure}[t]{1\textwidth}
    \centering
	\includegraphics[width=0.8\linewidth]{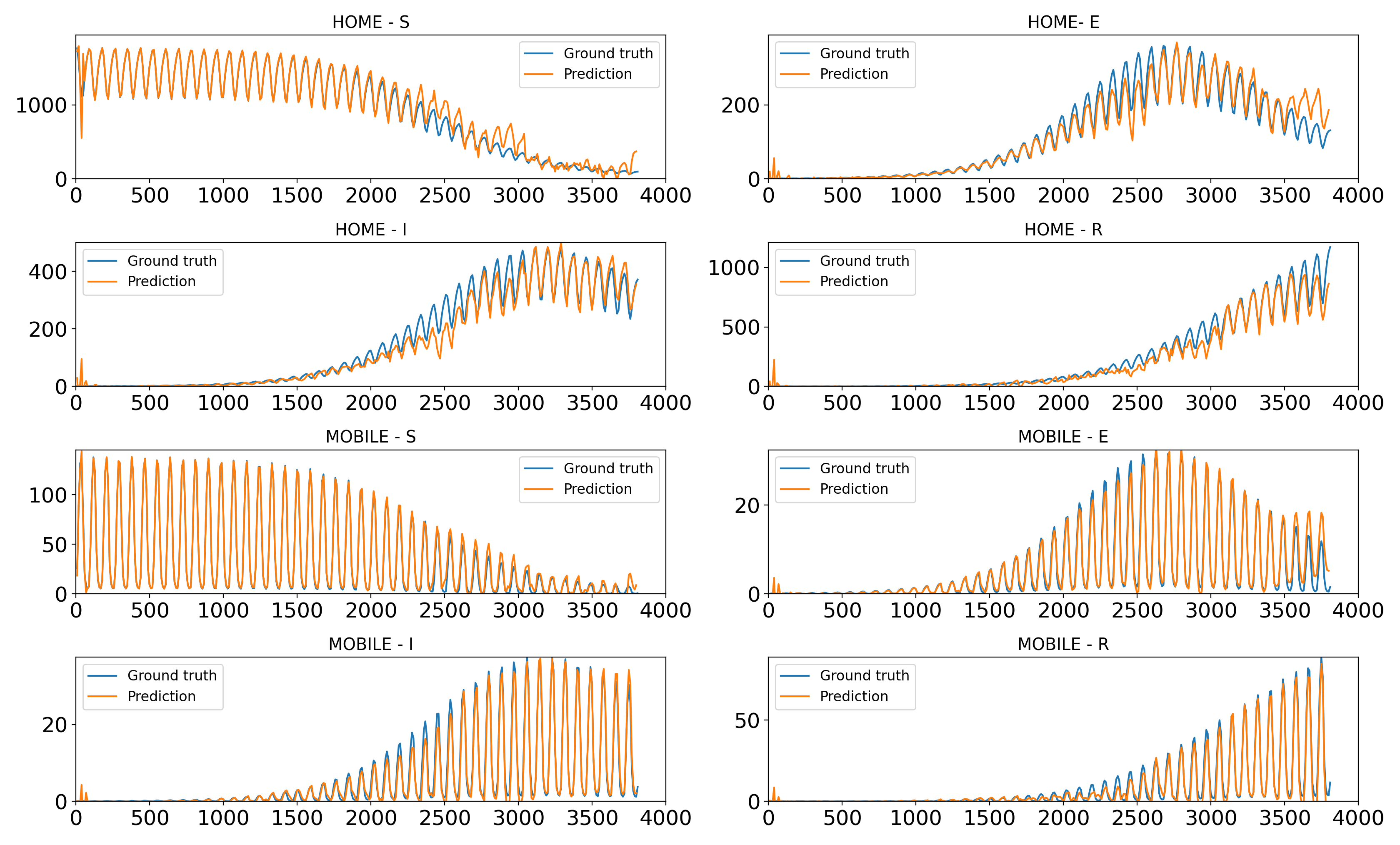}
	\caption{Starting at $9\times10^4$ seconds.}
	\label{fig:ffn0toend}
\end{subfigure}
\begin{subfigure}[t]{1\textwidth}
    \centering
	\includegraphics[width=0.8\linewidth]{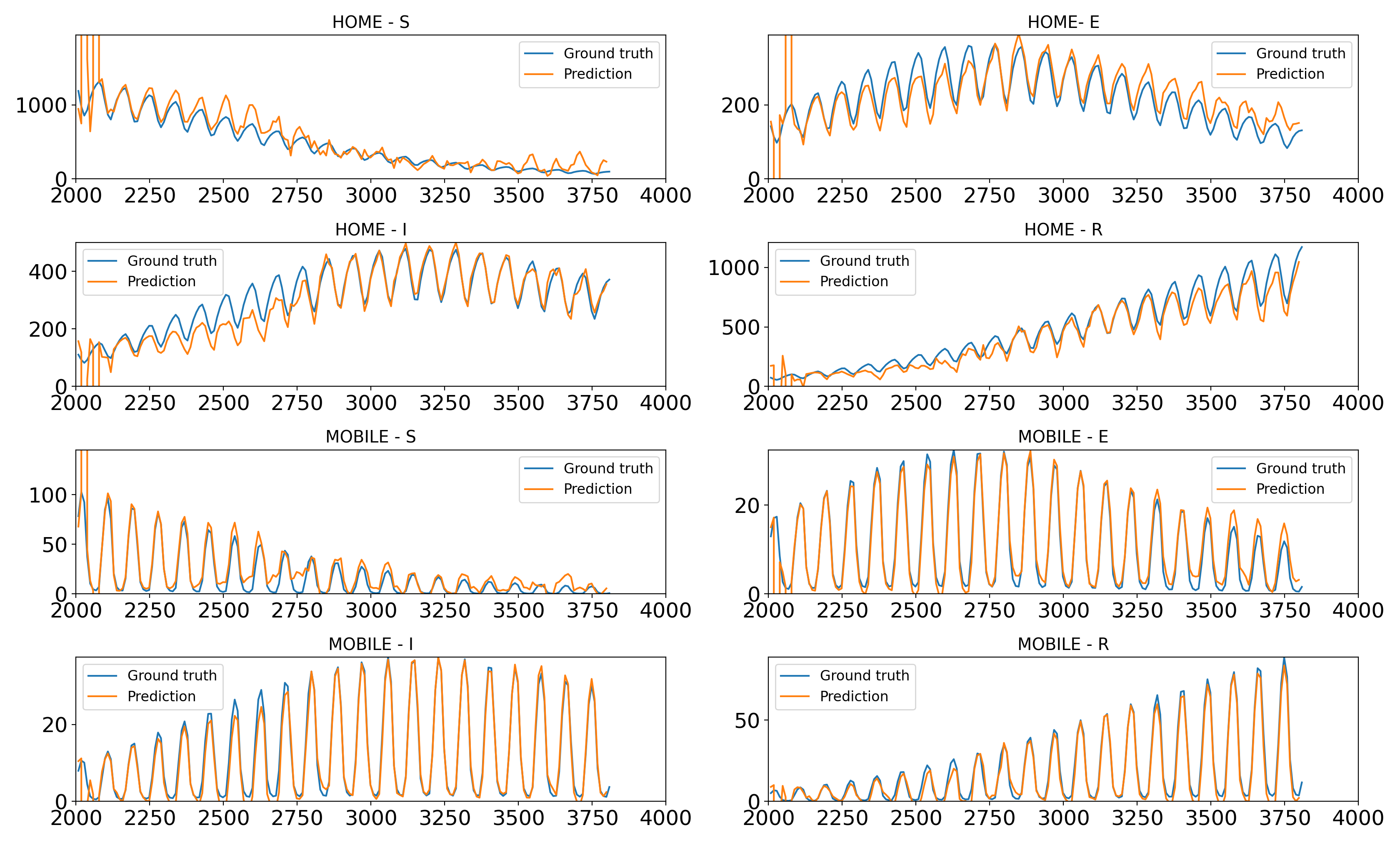}
	\caption{Starting at $2\times10^6$ seconds.}
	\label{fig:ddn2000toend}
\end{subfigure}
\caption{$f^{FFN}$ prediction (in number of people) at one point (marked as a red circle in Fig \ref{fig:regions}) of the domain over time starting from different time levels.} 
\label{fig:ffnPreds}
\end{figure*}

Similar to the previous models, Figure~\ref{fig:ffnPreds}a shows the prediction over time of $f^{FFN+BLUE}$ but in one point of the mesh (bottom-right corner of region 2 shown in Figure~\ref{fig:regions}). Each cycle in the curves corresponds to a period of one day. The process is repeated this time with the simulation starting at time step $2\times10^3$ ($2\times10^6$ seconds). The result over time for one point of the mesh (bottom-right corner of region 2) is presented in Figure~\ref{fig:ffnPreds}b.

Whilst the performance on the first batch of data is adequate and comparable to $f^{BDLSTM+BLUE}$ and $f^{PredictiveGAN}$, $f^{FFN+BLUE}$ fails when dealing with initial predictions of the second batch of data (starting at $2\times10^{6}$), thus being outperformed by the other two networks, with a RMSE of $9.535\times10^{9}$ in all Home compartments and $3.451\times10^{8}$ for Mobile compartments. In comparison, $f^{PredictiveGAN}$ has a RMSE of $3.285\times 10^{1}$ and $2.807\times 10^0$ for Home and Mobile compartments, respectively. And $f^{BDLSTM+BLUE}$ shows a RMSE of $3.049\times 10^1$ and $1.329\times10^0$ for Home and Mobile compartments, respectively. All stated RMSE are in number of people. Therefore, the final comparison of this paper only involves $f^{BDLSTM+BLUE}$ and $f^{PredictiveGAN}$

\subsection{Comparison between BDLSTM and predictive GAN}

Formatted as Jupyter notebooks, the codes for both experiments presented in this paper are publicly available at \url{https://github.com/c-quilo/SEIR-BDLSTM} (for the BDLSTM) and \url{https://github.com/viluiz/gan/tree/master/PredGAN} (for the GAN). The dependencies of the codes are Python (version 3.7), Numpy (version 1.18.5), Keras (version 2.4.3) and TensorFlow (version 2.4.0).
The final hyperparameters used in the Bidirectional Long Short-Term Memory and predictive GAN networks are given in Table~\ref{table-hyperparameters}. 

\begin{table*}[htbp]
\addtolength{\tabcolsep}{6pt}
\centering
\begin{tabular}{lll}
\hline \hline
& BDLSTM & GAN\\
\hline \hline
Epochs  & 500  & 55,000 \\
Batch size  & 32 & 256 \\
Hidden nodes  & 64  & n/a \\
Latent space size  & n/a  & 100 \\
Batch normalisation  & - & \checkmark (generator) \\
Layer normalisation & \checkmark & - \\
Dropout  & 0.5 & 0.3  (discriminator) \\
Activation function  & sigmoid$^{\,\dagger}$ & LeakyReLU (0.3$^{\,\ddagger}$) \\
Loss function   & Mean Square Error  &  Binary cross entropy  \\
Optimiser  & Nadam$^{\,\dagger\dagger}$  & Adam  \\
Learning rate   & 0.001   & 0.001   \\
$\beta_1$    & 0.9  & 0.9 \\
$\beta_2$    & 0.999   & 0.999 \\
$\epsilon$   & $10^{-7}$ & n/a \\ 
\hline \hline
\end{tabular}
\caption{Hyperparameters used for the data-corrected bidirectional LSTM and the predictive GAN. (${}^{\dagger\,}$Time distributed dense output layer with a sigmoid activation function, ${}^{\ddagger\,}$negative slope coefficient, ${}^{\dagger\dagger\,}$Adam with Nesterov momentum.)}
\label{table-hyperparameters}
\end{table*}

%
%



\begin{figure*}[t]
    \begin{subfigure}[t]{0.45\textwidth}
        \centering
        \includegraphics[width = 1\textwidth]{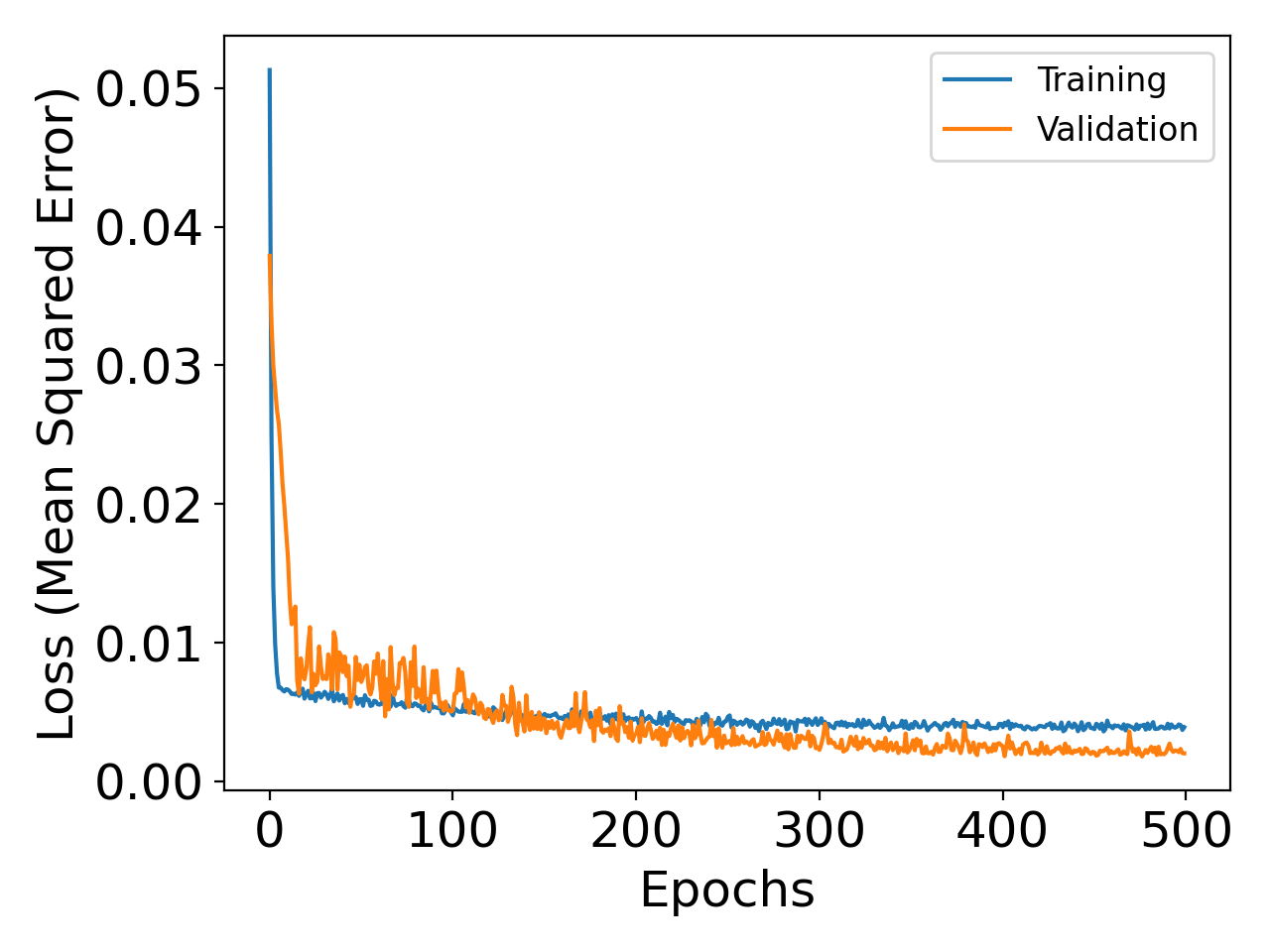}
        \caption{BDLSTM}
        \label{lossesbdlstm}
    \end{subfigure}
    \begin{subfigure}[t]{0.45\textwidth}
        \centering
        \includegraphics[width = 1\textwidth]{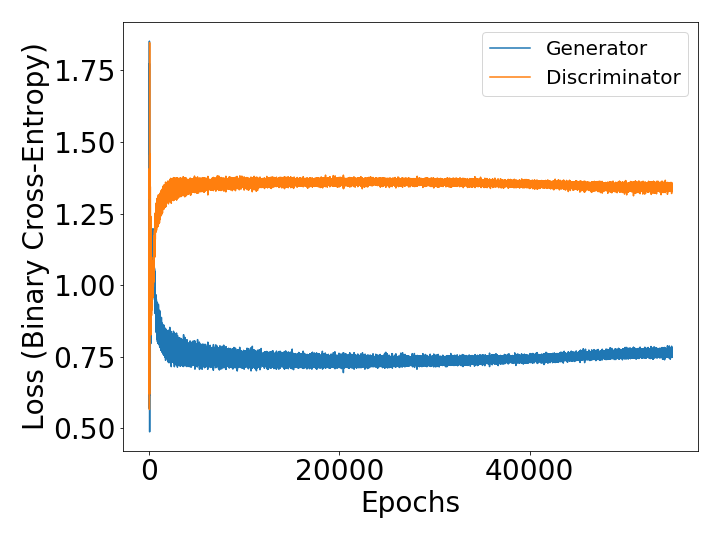}
        \caption{Predictive GAN}
        \label{lossespredgan}
    \end{subfigure}
\caption{Training losses of $f^{BDLSTM}$ (mean squared error), and the generator $G$ and discriminator $D$ (binary cross-entropy).}
\label{fig:traininglosses}
\end{figure*}

The training losses of both experiments, BDLSTM and GAN, are depicted in Fig \ref{fig:traininglosses}. 

\begin{figure*}[ht]
	\centering
	\includegraphics[width=1.0\linewidth]{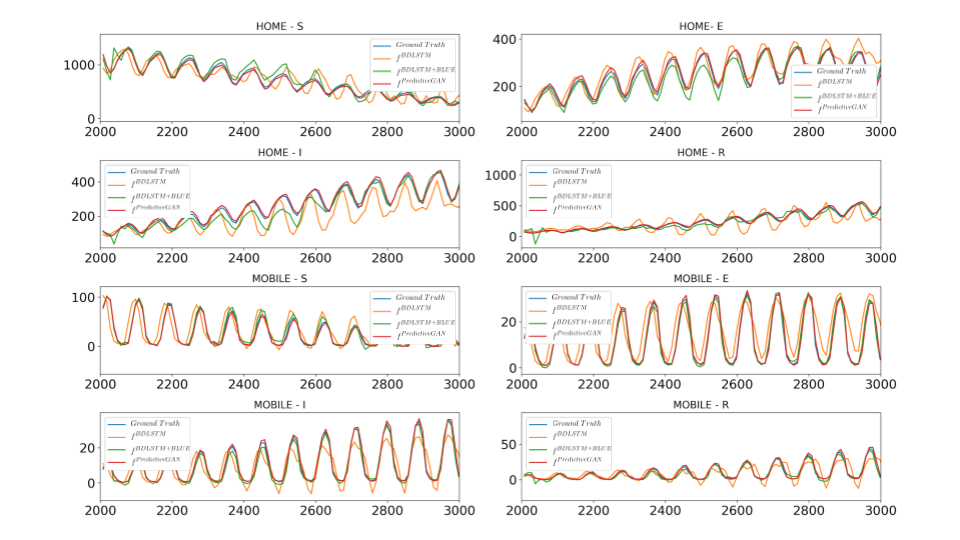}
	\caption{Comparison of forecasts (in number of people) produced by three methods: $f^{BDLSTM}$(orange), $f^{BDLSTM+BLUE}$ (green), and $f^{PredictiveGAN}$ (red), over time to the ground truth (blue). The forecast starts from t=2000 ($2 \times 10^6$ seconds) of the SEIRS model solution.}
	\label{fig:comparison}
\end{figure*}

Figure \ref{fig:comparison} presents a comparison over a short period of time (50 time-steps) including $f^{BDLSTM}$, the $f^{BDLSTM+BLUE}$, and $f^{PredictiveGAN}$. The BDLSTM benefits greatly from the data-correction with the BLUE estimator. However, it needs constant input from the model solution data to correct its trajectory. While the predictive GAN replicates the dynamics of the SEIRS model solution well, just with the input of 8 time levels at the start. Thus, $f^{PredictiveGAN}$ does not constantly look at the extended SEIRS model solution data.

\begin{figure*}[t]
\centering
    \begin{subfigure}[t]{0.48\textwidth}
        \includegraphics[width = 1\textwidth]{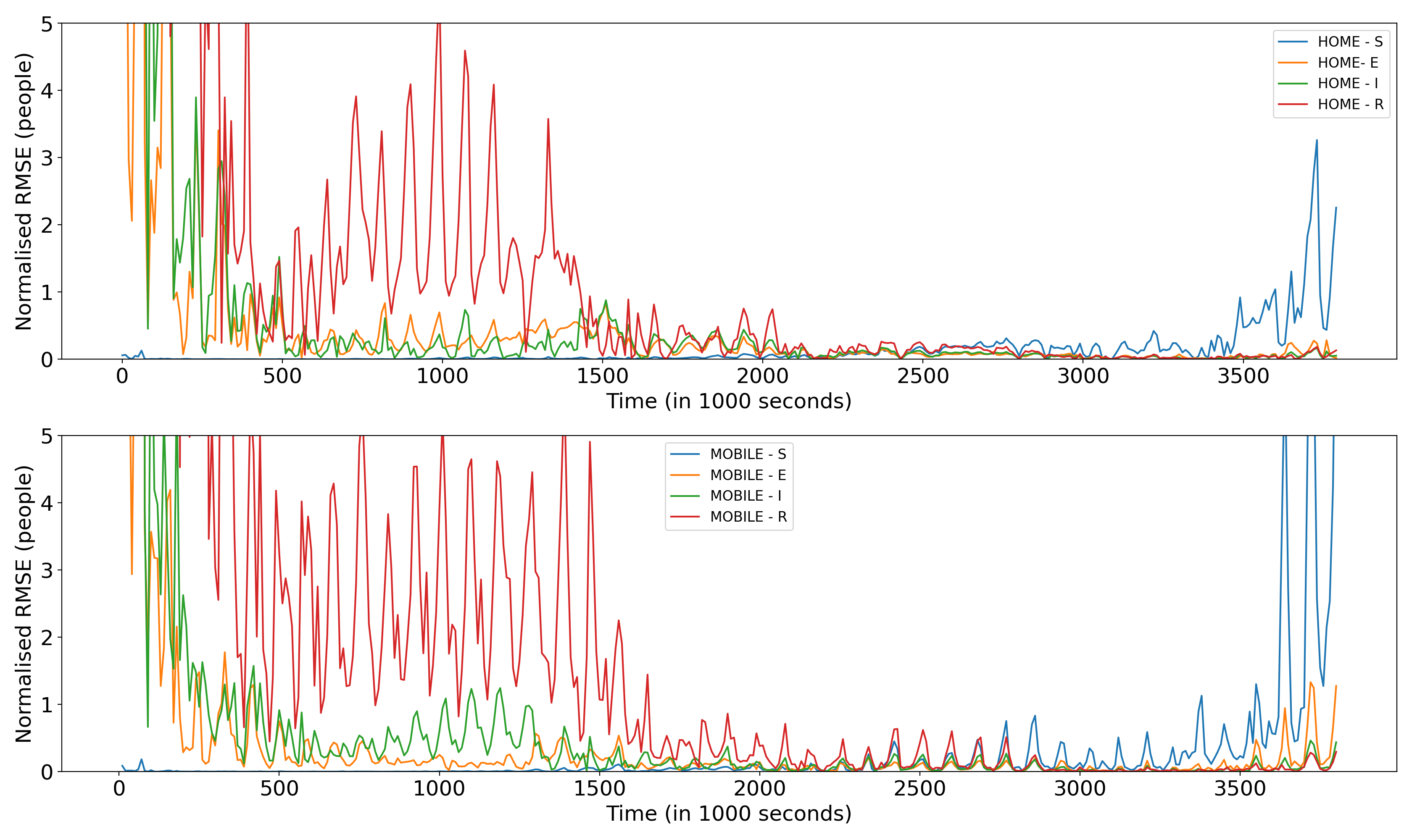}
        \label{NRMSEBDLSTM}
        \caption{$f^{BDLSTM+BLUE}$}
    \end{subfigure}
    \begin{subfigure}[t]{0.48\textwidth}
        \includegraphics[width = 1\textwidth]{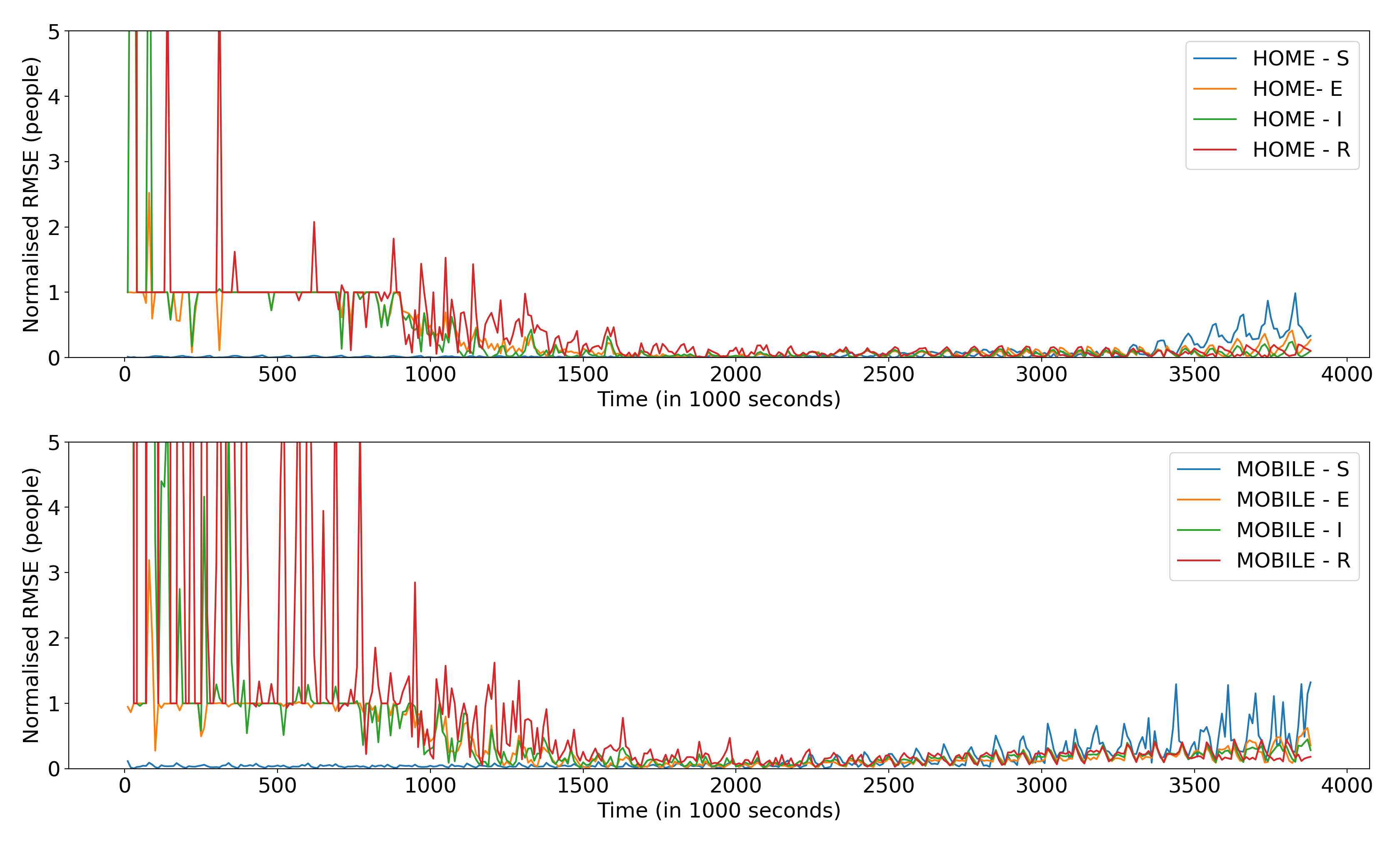}
        \label{NRMSEGAN}

        \caption{$f^{PredictiveGAN}$}
    \end{subfigure}
\caption{Time-series of the Normalised root mean squared error of the predictions for the Home (top) and Mobile (bottom) compartments. Left: $f^{BDLSTM+BLUE}$, Right: $f^{PredictiveGAN}$.}
\label{fig:RMSEBoth}
\end{figure*}

Figure \ref{fig:RMSEBoth} shows the normalised root mean squared error (NRMSE) over time for both digital twins. The mean was calculated using only the active regions of each compartment and group (Figure \ref{fig:regions}), i.e~the Home group is only considered in region 2, while the Mobile group is considered across the entire active region (all regions but 1). For this simulation, we start the prediction at time step $90$ ($9\times 10^{40}$ seconds).

The RMSE at time level $k$ 
is defined as the following:
\begin{equation}
RMSE^k =
\frac{\norm{\mathbf{u}^k - \mathbf{v}^k}_{2}}{\sqrt{m}}
\end{equation}
where $k$ is the time level, 
$\mathbf{u}^{k} \in \mathbb{R}^{m}$ are the predictions for a particular compartment and group, based on $f^{BDLSTM+BLUE}$ or $f^{PredictiveGAN}$ at time level $k$ (having mapped the output of the network back to the control-volume grid), $\mathbf{v}^k \in \mathbb{R}^{m}$ is the data from the extended SEIRS model solutions at time level $k$, $m$ is the number of active control volumes per compartment and group, and $\norm{}_{2}$ represents the Euclidean norm. A RMSE value is computed for the eight combinations of compartments and groups. The normalised RMSE at time level $k$ is defined by:
\begin{equation}
NRMSE^{k} = \frac{\norm{\mathbf{u}^k-\mathbf{v}^k}_2}{\norm{\mathbf{v}^k}_2}\ .
\end{equation}

In the prediction of the Home compartments using the $f^{BDLSTM+BLUE}$ prediction, it is worth noting that there is a decreasing trend of the Home - Recovered and Home - Infectious people, while the number of people in Home - Susceptible increases towards the end of the dataset surpassing the normalised RMSE of the other compartments and groups. The predictions by $f^{PredictiveGAN}$ on the the Home group present similar behaviour over time. However, the decreasing trends are more rapid and the increased error of the Home - Susceptible compartment is smaller towards the end of the dataset.

There is a very similar behaviour for the Mobile groups in both the BDLSTM and GAN predictions. There is a decreasing trend for the Mobile - Exposed, Mobile - Infectious and Mobile - Recovered people for both experiments. Additionally, the error seen for people in  Home - Susceptible increases over time in both experiments. A summary of the average normalised RMSE over time is shown in Table \ref{table1}.

\begin{table*}[ht]
\centering
  \caption{Average normalised RMSE over time for both $f^{BDLSTM+BLUE}$, and $f^{PredictiveGAN}$ over the 4 compartments and 2 groups. The average does not consider the first 50 time-steps as the normalised RMSE is too sensitive during this period.}
  \label{table1}
  \begin{tabular}{|l|r|r|r|r|r|r|r|r|}
    \hline
    
    & H-S & H-E & H-I & H-R & M-S & M-E & M-I & M-R\\
    \hline
    $f^{BDLSTM+BLUE}$ & 0.179 & 0.170 & 0.164 & 0.888 & 0.409 & 0.176 & 0.192 & 0.353\\
    $f^{PredictiveGAN}$ & 0.078 & 0.210 & 0.182 & 0.264 & 0.175 & 0.281 & 0.287 & 0.503\\
    \hline
    \end{tabular}
\end{table*}

\begin{figure*}[t]
	\centering
	\includegraphics[trim = {1cm 1cm 1cm 1cm},clip,width=1\linewidth]{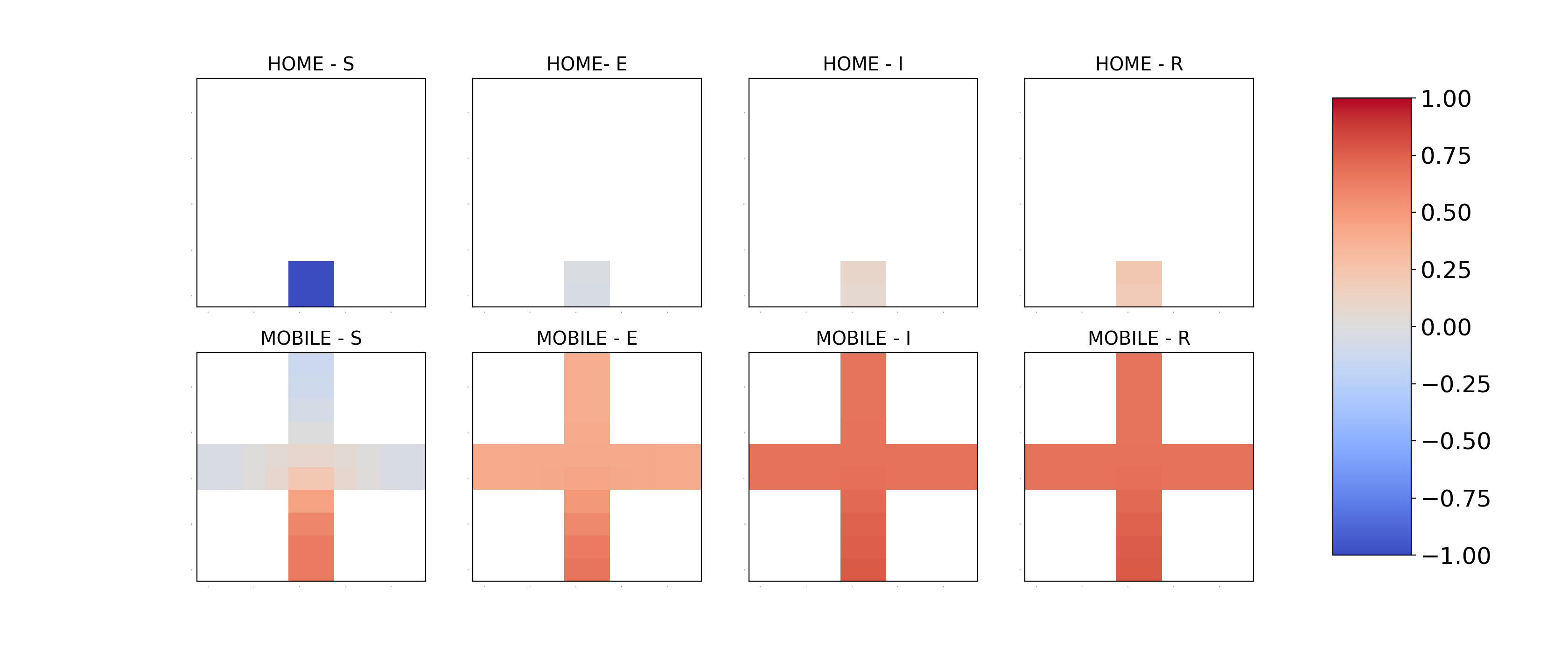}
	\caption{Spatial skill score over the mesh for all 4 compartments and 2 groups. If the skill score is less than zero, $f^{PredictiveGAN}$ has more skill at predicting that region. Otherwise, if the skill score is greater than 0, the $f^{BDLSTM+BLUE}$ is better at predicting that region. The first 50 time-steps were not considered.}
	\label{fig:ss}
\end{figure*}

In order to compare the skill of the $f^{BDLSTM+BLUE}$ and $f^{PredictiveGAN}$, we look at the spatial skill score (SS):
\begin{equation}
SS = 1 - \frac{RMSE_{f^{BDLSTM+BLUE}}}{RMSE_{f^{PredictiveGAN}}}
\end{equation}
where $RMSE_{f^{BDLSTM+BLUE}}$ and $RMSE_{f^{PredictiveGAN}}$ are the spatial RMSE averaged over time on each region. The spatial SS is depicted in Figure~\ref{fig:ss}. If $SS<0$, the predictive GAN has more skill at predicting that region. Otherwise, if $SS>0$, the $f^{BDLSTM+BLUE}$ is better at predicting that region. While $f^{PredictiveGAN}$ outperforms $f^{BDLSTM+BLUE}$ for the prediction of the Home group (compartments S, E, I and R), in general, the data-corrected BDLSTM produces more accurate predictions for the Mobile - Infectious and Mobile - Recovered people.

The execution times with optimisation for both experiments are shown in Table \ref{table:times}. These execution times are concerning a set of 9 time-steps. The speed-up for the original simulation is also shown. 
If optimisation is included, the $f^{BDLSTM+BLUE}$ prediction is 2 orders of magnitude faster than $f^{PredictiveGAN}$. 

\begin{table*}[ht]
\centering
  \caption{Execution times with optimisation of a single set of 9 time-steps, and the speed-up of each method with respect to the original simulation. The original simulation does not include an optimisation, thus both speed-up times are with respect to the simulation execution time for 9 time-steps.}
  \label{table:times}
  \begin{tabular}{|l|c|c|c|}
    \hline
    & \multicolumn{1}{c|}{Execution times (s)} & \multicolumn{1}{c|}{Speed-up (-)}& Storage size\\
    \hline
     SEIRS & \multicolumn{1}{c|}{0.45} & \multicolumn{1}{c|}{-}&7.25 Mb\\
    \hline
    \hline
    $f^{BDLSTM+BLUE}$ & $1.6 \times 10^{-2}$ & 28.12 & 0.14 Mb\\
    $f^{PredictiveGAN}$ & $1.9 \times 10^0$ & 0.24 & 0.14 Mb\\
    \hline
    \end{tabular}
\end{table*}

\section{Discussion}
\label{sec:discussion}
These experiments serve as a proof of concept for digital twins of SEIRS models. The predictions produced by the predictive GAN outperform the data-corrected BDLSTM in the Susceptible compartments in both Home and Mobile groups, while the data-corrected prediction of the BDLSTM outperforms the predictive GAN in the Exposed, Infectious and Recovered compartments. However, it is important to note that the predictions produced by the BDLSTM are data-corrected using the BLUE optimisation. The predictive GAN also includes optimisation, but it is capable to generalise over time just by optimising observational data at the beginning of its prediction.

\begin{itemize}
    \item The $f^{BDLSTM}$ provides fast forecasts which are up to 4 orders of magnitude faster than the simulation. However, it was observed that the BDLSTM diverges quickly from the model solution when the predicted output is used as an input to predict the following time-step.
    \item This was fixed by adding a data-correction step, using BLUE. The produced forecasts using this method are 2 orders of magnitude faster than the extended SEIRS model solution. However, it has the disadvantage of constantly having the extended SEIRS model solution as input to correct the trajectory of the forecast.
    \item While $f^{BDLSTM+BLUE}$ outperforms $f^{PredictiveGAN}$ at producing forecasts of the extended SEIRS model solution, $f^{PredictiveGAN}$ has the great advantage of not needing a constant stream of data from the extended SEIRS model. The $f^{PredictiveGAN}$ manages to predict the dynamics of the extended SEIRS model accurately with only the input of 8 time-steps at the start of the simulation. These 8 time-steps serve as a constraint to initialise the forecast of $f^{PredictiveGAN}$. Additionally, GANs can generate reliable information from random noise, which LSTMs are not designed to do. Nonetheless, the execution times of $f^{PredictiveGAN}$ are slower than those of the $f^{BDLSTM+BLUE}$ by 2 orders of magnitude.
    \item $f^{PredictiveGAN}$ has great potential when applied in larger problems. In any case, for a more demanding SEIRS model (with more compartments or with a higher spatial resolution for example), the speed-ups of both digital twins are expected to improve.
\end{itemize}

Therefore, a combination of both techniques will be valuable in the future for a more accurate prediction that includes information from the time series, using an LSTM, and creating realistic information trained with adversarial networks. Similar efforts in combining LSTM and GAN/adversarial training have been studied for Electrocardiograms \citep{zhu2019electrocardiogram} and classical music generation \citep{mogren2016c}. Thus, the prediction of future time-steps will be embedded into the GAN, without requiring further optimisation to make a prediction. This method will diminish execution times, with the caveat that training GANs come at a higher computational cost. Nonetheless, an application on SEIRS modelling and more specifically applied to COVID-19 have not been implemented.

Our choice of using a BDLSTM is supported by previous studies using BDLSTMs for COVID-19 prediction. In \citet{shahid2020predictions}, the authors show a comparison of different deep learning methods for forecasting COVID-19 time series data and concluded that a BDLSTM shows robustness and it is an appropriate predictor for this type of data, outperforming a vanilla-LSTM and a Gated Recurrent Unit network. \citet{chatterjee2020statistical} also presents that a BDLSTM is a strong predictive model for forecasting new cases and resulting deaths of COVID-19. Furthermore, the comparison of our two models, $f^{BDLSTM+BLUE}$ and $f^{PredictiveGAN}$, against a feed-forward network show how the former models, as they take advantage of temporal information, outperform the latter.

\section{Conclusions and future work}
\label{section:conclusions}

In this paper, we have presented two methods for creating digital twins of a SEIRS model with four compartments and two groups. The SEIRS model has been extended to be able to model both the spatial and temporal spread of the virus. The digital twins or non-intrusive reduced-order models (NIROMS) were used for predicting the future states of the model comparing the evolution of these experiments to the ground truth. The first experiment uses a Bidirectional Long Short-term memory network (BDLSTM), while the second experiment utilises a Predictive Generative Adversarial Network (GAN). The prediction produced by the predictive GAN outperforms the predictions by the data-corrected BDLSTM in the Susceptible compartments. Furthermore, GANs can generate reliable information from random noise. This novel approach using data-corrected optimisation using GANs shows very promising results for time-series prediction.

In summary, this paper proposed a novel Reduced Order Model (ROM) based on a Bidirectional Long Short-Term Memory (BDLSTM) network with a data-correction step derived from BLUE for improved accuracy. A novel predictive GAN is also used. Finally, we compared the two models. The novelty of this paper also relies on that this is the first time that reduced-order modelling techniques have been applied to virus modelling. 

Future work involves the combination of LSTM (unidirectional or bidirectional) with a GAN to produce more accurate forecasts that take advantage of the time-series information along with realistic predictions produced by the GAN. Additionally, these frameworks could be applied to larger domains of idealised towns including more compartments to study more realistic epidemiological models.

\printglossary[type=main,style=long,nonumberlist]

\section*{Acknowledgements}
 This work is supported by the EPSRC grant EP/T003189/1 Health assessment across biological length scales for personal pollution exposure and its mitigation (INHALE), grant EP/N010221/1 Managing Air for Green Inner Cities (MAGIC) consortium, the PREMIERE programme grant (EP/T000414/1), the MUFFINS grant (EP/P033180/1), the RELIANT grant (EP/V036777/1), and by BECAS CHILE, a governmental Chilean scholarship from the Agencia Nacional de Investigaci\'on y Desarrollo (ANID). This work has been undertaken, in part, as a contribution to `Rapid Assistance in Modelling the Pandemic' (RAMP), initiated by the Royal Society. In particular, we would like to acknowledge the useful discussion had within the Environmental and Aerosol Transmission group of RAMP, coordinated by Profs Paul Linden and Christopher Pain.





\clearpage
\bibliographystyle{model1-num-names}
\bibliography{references.bib}

\begin{thebibliography}{6}
\providecommand{\natexlab}[1]{#1}
\providecommand{\url}[1]{\texttt{#1}}
\expandafter\ifx\csname urlstyle\endcsname\relax
  \providecommand{\doi}[1]{doi: #1}\else
  \providecommand{\doi}{doi: \begingroup \urlstyle{rm}\Url}\fi

\bibitem[Abadi et~al.(2015)Abadi, Agarwal, Barham, Brevdo, Chen, Citro,
  Corrado, Davis, Dean, Devin, Ghemawat, Goodfellow, Harp, Irving, Isard, Jia,
  Jozefowicz, Kaiser, Kudlur, Levenberg, Man\'{e}, Monga, Moore, Murray, Olah,
  Schuster, Shlens, Steiner, Sutskever, Talwar, Tucker, Vanhoucke, Vasudevan,
  Vi\'{e}gas, Vinyals, Warden, Wattenberg, Wicke, Yu, and
  Zheng]{tensorflow:2015}
M.~Abadi, A.~Agarwal, P.~Barham, E.~Brevdo, Z.~Chen, C.~Citro, G.~S. Corrado,
  A.~Davis, J.~Dean, M.~Devin, S.~Ghemawat, I.~Goodfellow, A.~Harp, G.~Irving,
  M.~Isard, Y.~Jia, R.~Jozefowicz, L.~Kaiser, M.~Kudlur, J.~Levenberg,
  D.~Man\'{e}, R.~Monga, S.~Moore, D.~Murray, C.~Olah, M.~Schuster, J.~Shlens,
  B.~Steiner, I.~Sutskever, K.~Talwar, P.~Tucker, V.~Vanhoucke, V.~Vasudevan,
  F.~Vi\'{e}gas, O.~Vinyals, P.~Warden, M.~Wattenberg, M.~Wicke, Y.~Yu, and
  X.~Zheng.
\newblock {TensorFlow}: Large-scale machine learning on heterogeneous systems,
  2015.
\newblock URL \url{https://www.tensorflow.org/}.
\newblock Software available from tensorflow.org.

\bibitem[Baydin et~al.(2017)Baydin, Pearlmutter, Radul, and Siskind]{baydin:17}
A.~G. Baydin, B.~A. Pearlmutter, A.~A. Radul, and J.~M. Siskind.
\newblock Automatic differentiation in machine learning: a survey.
\newblock \emph{The Journal of Machine Learning Research}, 18\penalty0
  (1):\penalty0 5595--5637, 2017.

\bibitem[Goodfellow et~al.(2014)Goodfellow, Pouget-Abadie, Mirza, Xu,
  Warde-Farley, Ozair, Courville, and Bengio]{goodfellow:14}
I.~Goodfellow, J.~Pouget-Abadie, M.~Mirza, B.~Xu, D.~Warde-Farley, S.~Ozair,
  A.~Courville, and Y.~Bengio.
\newblock Generative adversarial nets.
\newblock In \emph{Advances in neural information processing systems}, pages
  2672--2680, 2014.

\bibitem[Linnainmaa(1976)]{linnainmaa:76}
S.~Linnainmaa.
\newblock Taylor expansion of the accumulated rounding error.
\newblock \emph{BIT Numerical Mathematics}, 16\penalty0 (2):\penalty0 146--160,
  1976.

\bibitem[Radford et~al.(2015)Radford, Metz, and Chintala]{radford:15}
A.~Radford, L.~Metz, and S.~Chintala.
\newblock Unsupervised representation learning with deep convolutional
  generative adversarial networks.
\newblock \emph{arXiv preprint arXiv:1511.06434}, 2015.

\bibitem[Wengert(1964)]{wengert:64}
R.~E. Wengert.
\newblock A simple automatic derivative evaluation program.
\newblock \emph{Communications of the ACM}, 7\penalty0 (8):\penalty0 463--464,
  1964.

\end{thebibliography}


\begin{thebibliography}{75}
\expandafter\ifx\csname natexlab\endcsname\relax\def\natexlab#1{#1}\fi
\providecommand{\bibinfo}[2]{#2}
\ifx\xfnm\relax \def\xfnm[#1]{\unskip,\space#1}\fi
\bibitem[{Dong et~al.(2020)Dong, Du, and Gardner}]{dong2020interactive}
\bibinfo{author}{E.~Dong}, \bibinfo{author}{H.~Du},
  \bibinfo{author}{L.~Gardner},
\newblock \bibinfo{title}{{An interactive web-based dashboard to track COVID-19
  in real time}},
\newblock \bibinfo{journal}{The Lancet infectious diseases}
  \bibinfo{volume}{20} (\bibinfo{year}{2020}) \bibinfo{pages}{533--534}.
\bibitem[{Park et~al.(2020)Park, Cook, Lim, Sun, and
  Dickens}]{park2020systematic}
\bibinfo{author}{M.~Park}, \bibinfo{author}{A.~R. Cook}, \bibinfo{author}{J.~T.
  Lim}, \bibinfo{author}{Y.~Sun}, \bibinfo{author}{B.~L. Dickens},
\newblock \bibinfo{title}{{A systematic review of COVID-19 epidemiology based
  on current evidence}},
\newblock \bibinfo{journal}{Journal of Clinical Medicine} \bibinfo{volume}{9}
  (\bibinfo{year}{2020}) \bibinfo{pages}{967}.
\bibitem[{Auchincloss and Diez~Roux(2008)}]{auchincloss2008new}
\bibinfo{author}{A.~H. Auchincloss}, \bibinfo{author}{A.~V. Diez~Roux},
\newblock \bibinfo{title}{A new tool for epidemiology: the usefulness of
  dynamic-agent models in understanding place effects on health},
\newblock \bibinfo{journal}{American journal of epidemiology}
  \bibinfo{volume}{168} (\bibinfo{year}{2008}) \bibinfo{pages}{1--8}.
\bibitem[{Cuevas(2020)}]{cuevas2020agent}
\bibinfo{author}{E.~Cuevas},
\newblock \bibinfo{title}{{An agent-based model to evaluate the COVID-19
  transmission risks in facilities}},
\newblock \bibinfo{journal}{Computers in biology and medicine}
  \bibinfo{volume}{121} (\bibinfo{year}{2020}) \bibinfo{pages}{103827}.
\bibitem[{Shamil et~al.(2021)Shamil, Farheen, Ibtehaz, Khan, and
  Rahman}]{shamil2021agent}
\bibinfo{author}{M.~S. Shamil}, \bibinfo{author}{F.~Farheen},
  \bibinfo{author}{N.~Ibtehaz}, \bibinfo{author}{I.~M. Khan},
  \bibinfo{author}{M.~S. Rahman},
\newblock \bibinfo{title}{{An Agent-Based Modeling of COVID-19: Validation,
  Analysis, and Recommendations}},
\newblock \bibinfo{journal}{Cognitive Computation}  (\bibinfo{year}{2021})
  \bibinfo{pages}{1--12}.
\bibitem[{Li and Muldowney(1995)}]{li1995global}
\bibinfo{author}{M.~Y. Li}, \bibinfo{author}{J.~S. Muldowney},
\newblock \bibinfo{title}{{Global stability for the SEIR model in
  epidemiology}},
\newblock \bibinfo{journal}{Mathematical biosciences} \bibinfo{volume}{125}
  (\bibinfo{year}{1995}) \bibinfo{pages}{155--164}.
\bibitem[{R\u{a}dulescu et~al.(2020)R\u{a}dulescu, Williams, and
  Cavanagh}]{Radulescu2020}
\bibinfo{author}{A.~R\u{a}dulescu}, \bibinfo{author}{C.~Williams},
  \bibinfo{author}{K.~Cavanagh},
\newblock \bibinfo{title}{Management strategies in a {SEIR}-type model of
  {COVID} 19 community spread},
\newblock \bibinfo{journal}{Scientific Reports} \bibinfo{volume}{10}
  (\bibinfo{year}{2020}) \bibinfo{pages}{21256}.
\bibitem[{Basu and Andrews(2013)}]{basu2013complexity}
\bibinfo{author}{S.~Basu}, \bibinfo{author}{J.~Andrews},
\newblock \bibinfo{title}{Complexity in mathematical models of public health
  policies: a guide for consumers of models},
\newblock \bibinfo{journal}{PLoS Med} \bibinfo{volume}{10}
  (\bibinfo{year}{2013}) \bibinfo{pages}{e1001540}.
\bibitem[{Rock et~al.(2014)Rock, Brand, Moir, and Keeling}]{rock2014dynamics}
\bibinfo{author}{K.~Rock}, \bibinfo{author}{S.~Brand},
  \bibinfo{author}{J.~Moir}, \bibinfo{author}{M.~J. Keeling},
\newblock \bibinfo{title}{Dynamics of infectious diseases},
\newblock \bibinfo{journal}{Reports on Progress in Physics}
  \bibinfo{volume}{77} (\bibinfo{year}{2014}) \bibinfo{pages}{026602}.
\bibitem[{Cameron et~al.(2004)Cameron, Lyons, and
  Kenworthy}]{cameron2004trends}
\bibinfo{author}{I.~Cameron}, \bibinfo{author}{T.~Lyons},
  \bibinfo{author}{J.~Kenworthy},
\newblock \bibinfo{title}{Trends in vehicle kilometres of travel in world
  cities, 1960--1990: underlying drivers and policy responses},
\newblock \bibinfo{journal}{Transport policy} \bibinfo{volume}{11}
  (\bibinfo{year}{2004}) \bibinfo{pages}{287--298}.
\bibitem[{Pavlidis et~al.(2014)Pavlidis, Xie, Percival, Gomes, Pain, and
  Matar}]{pavlidis:14}
\bibinfo{author}{D.~Pavlidis}, \bibinfo{author}{Z.~Xie}, \bibinfo{author}{J.~R.
  Percival}, \bibinfo{author}{J.~L. Gomes}, \bibinfo{author}{C.~C. Pain},
  \bibinfo{author}{O.~K. Matar},
\newblock \bibinfo{title}{{Two-and three-phase horizontal slug flow simulations
  using an interface-capturing compositional approach}},
\newblock \bibinfo{journal}{International Journal of Multiphase Flow}
  \bibinfo{volume}{67} (\bibinfo{year}{2014}) \bibinfo{pages}{85--91}.
\bibitem[{Eubank(2002)}]{eubank2002scalable}
\bibinfo{author}{S.~Eubank},
\newblock \bibinfo{title}{Scalable, efficient epidemiological simulation},
\newblock in: \bibinfo{booktitle}{Proceedings of the 2002 ACM symposium on
  Applied computing}, pp. \bibinfo{pages}{139--145}.
\bibitem[{Cooke and Van Den~Driessche(1996)}]{cooke1996analysis}
\bibinfo{author}{K.~L. Cooke}, \bibinfo{author}{P.~Van Den~Driessche},
\newblock \bibinfo{title}{Analysis of an seirs epidemic model with two delays},
\newblock \bibinfo{journal}{Journal of Mathematical Biology}
  \bibinfo{volume}{35} (\bibinfo{year}{1996}) \bibinfo{pages}{240--260}.
\bibitem[{Song et~al.(2019)Song, Lou, and Xiao}]{song2019spatial}
\bibinfo{author}{P.~Song}, \bibinfo{author}{Y.~Lou}, \bibinfo{author}{Y.~Xiao},
\newblock \bibinfo{title}{A spatial {SEIRS} reaction-diffusion model in
  heterogeneous environment},
\newblock \bibinfo{journal}{Journal of Differential Equations}
  \bibinfo{volume}{267} (\bibinfo{year}{2019}) \bibinfo{pages}{5084--5114}.
\bibitem[{Xiao et~al.(2015)Xiao, Fang, Buchan, Pain, Navon, and
  Muggeridge}]{xiao2015non}
\bibinfo{author}{D.~Xiao}, \bibinfo{author}{F.~Fang},
  \bibinfo{author}{A.~Buchan}, \bibinfo{author}{C.~Pain},
  \bibinfo{author}{I.~Navon}, \bibinfo{author}{A.~Muggeridge},
\newblock \bibinfo{title}{{Non-intrusive reduced order modelling of the
  Navier--Stokes equations}},
\newblock \bibinfo{journal}{Computer Methods in Applied Mechanics and
  Engineering} \bibinfo{volume}{293} (\bibinfo{year}{2015})
  \bibinfo{pages}{522--541}.
\bibitem[{Hesthaven and Ubbiali(2018)}]{hesthaven2018non}
\bibinfo{author}{J.~S. Hesthaven}, \bibinfo{author}{S.~Ubbiali},
\newblock \bibinfo{title}{{Non-intrusive reduced order modeling of nonlinear
  problems using neural networks}},
\newblock \bibinfo{journal}{Journal of Computational Physics}
  \bibinfo{volume}{363} (\bibinfo{year}{2018}) \bibinfo{pages}{55--78}.
\bibitem[{Quilodr{\'a}n~Casas(2018)}]{quilodran2018fast}
\bibinfo{author}{C.~A. Quilodr{\'a}n~Casas}, \bibinfo{title}{Fast ocean data
  assimilation and forecasting using a neural-network reduced-space regional
  ocean model of the north Brazil current}, Ph.D. thesis, Imperial College
  London, \bibinfo{year}{2018}.
\bibitem[{Quilodr{\'a}n~Casas et~al.(2020)Quilodr{\'a}n~Casas, Arcucci, Wu,
  Pain, and Guo}]{casas2020reduced}
\bibinfo{author}{C.~Quilodr{\'a}n~Casas}, \bibinfo{author}{R.~Arcucci},
  \bibinfo{author}{P.~Wu}, \bibinfo{author}{C.~Pain}, \bibinfo{author}{Y.-K.
  Guo},
\newblock \bibinfo{title}{A reduced order deep data assimilation model},
\newblock \bibinfo{journal}{Physica D: Nonlinear Phenomena}
  (\bibinfo{year}{2020}) \bibinfo{pages}{132615}.
\bibitem[{Lever et~al.(2017)Lever, Krzywinski, and Altman}]{lever2017points}
\bibinfo{author}{J.~Lever}, \bibinfo{author}{M.~Krzywinski},
  \bibinfo{author}{N.~Altman}, \bibinfo{title}{{Points of significance:
  Principal component analysis}}, \bibinfo{year}{2017}.
\bibitem[{Quilodr{\'a}n~Casas et~al.(2020)Quilodr{\'a}n~Casas, Arcucci, and
  Guo}]{casas2020urban}
\bibinfo{author}{C.~Quilodr{\'a}n~Casas}, \bibinfo{author}{R.~Arcucci},
  \bibinfo{author}{Y.~Guo},
\newblock \bibinfo{title}{Urban air pollution forecasts generated from latent
  space representations},
\newblock in: \bibinfo{booktitle}{ICLR 2020 Workshop on Integration of Deep
  Neural Models and Differential Equations}.
\bibitem[{Quilodr{\'a}n-Casas et~al.(2021)Quilodr{\'a}n-Casas, Arcucci, Mottet,
  Guo, and Pain}]{quilodran2021adversarial}
\bibinfo{author}{C.~Quilodr{\'a}n-Casas}, \bibinfo{author}{R.~Arcucci},
  \bibinfo{author}{L.~Mottet}, \bibinfo{author}{Y.~Guo},
  \bibinfo{author}{C.~Pain},
\newblock \bibinfo{title}{Adversarial autoencoders and adversarial lstm for
  improved forecasts of urban air pollution simulations},
\newblock \bibinfo{journal}{Published as a workshop paper at ICLR 2021 SimDL
  Workshop}  (\bibinfo{year}{2021}).
\bibitem[{Phillips et~al.(2021)Phillips, Heaney, Smith, and
  Pain}]{phillips2020autoencoder}
\bibinfo{author}{T.~R.~F. Phillips}, \bibinfo{author}{C.~E. Heaney},
  \bibinfo{author}{P.~N. Smith}, \bibinfo{author}{C.~C. Pain},
\newblock \bibinfo{title}{An autoencoder-based reduced-order model for
  eigenvalue problems with application to neutron diffusion},
\newblock \bibinfo{journal}{International Journal for Numerical Methods in
  Engineering} \bibinfo{volume}{(accepted
  \href{https://doi.org/10.1002/nme.6681}{doi.org/10.1002/nme.6681})}
  (\bibinfo{year}{2021}).
\bibitem[{Bui-Thanh et~al.(2003)Bui-Thanh, Damodaran, and
  Willcox}]{Bui-Thanh2003}
\bibinfo{author}{T.~Bui-Thanh}, \bibinfo{author}{M.~Damodaran},
  \bibinfo{author}{K.~Willcox},
\newblock \bibinfo{title}{{Proper Orthogonal Decomposition Extensions for
  Parametric Applications in Compressible Aerodynamics}},
\newblock in: \bibinfo{booktitle}{21st AIAA Applied Aerodynamics Conference},
  \bibinfo{year}{2003}.
\bibitem[{Breitkopf et~al.(2017)Breitkopf, Lepot, Sainvitu, and
  Villon}]{Benamara2017}
\bibinfo{author}{P.~Breitkopf}, \bibinfo{author}{I.~Lepot},
  \bibinfo{author}{C.~Sainvitu}, \bibinfo{author}{P.~Villon},
\newblock \bibinfo{title}{Multi-fidelity {POD} surrogate-assisted optimization:
  {C}oncept and aero-design study},
\newblock \bibinfo{journal}{Structural and Multidisciplinary Optimization}
  \bibinfo{volume}{56} (\bibinfo{year}{2017}) \bibinfo{pages}{1387--1412}.
\bibitem[{Xiao et~al.(2019)Xiao, Fang, Heaney, Navon, and Pain}]{Xiao2019}
\bibinfo{author}{D.~Xiao}, \bibinfo{author}{F.~Fang}, \bibinfo{author}{C.~E.
  Heaney}, \bibinfo{author}{I.~M. Navon}, \bibinfo{author}{C.~C. Pain},
\newblock \bibinfo{title}{A domain decomposition method for the non-intrusive
  reduced order modelling of fluid flow},
\newblock \bibinfo{journal}{Computer Methods in Applied Mechanics and
  Engineering} \bibinfo{volume}{354} (\bibinfo{year}{2019})
  \bibinfo{pages}{307--330}.
\bibitem[{Aversano et~al.(2019)Aversano, Bellemans, Li, Coussement, Gicquel,
  and Parente}]{Aversano2019}
\bibinfo{author}{G.~Aversano}, \bibinfo{author}{A.~Bellemans},
  \bibinfo{author}{Z.~Li}, \bibinfo{author}{A.~Coussement},
  \bibinfo{author}{O.~Gicquel}, \bibinfo{author}{A.~Parente},
\newblock \bibinfo{title}{Application of reduced-order models based on {PCA} \&
  {K}riging for the development of digital twins of reacting flow
  applications},
\newblock \bibinfo{journal}{Computers \& Chemical Engineering}
  \bibinfo{volume}{121} (\bibinfo{year}{2019}) \bibinfo{pages}{422--441}.
\bibitem[{Kaiser et~al.(2014)Kaiser, Noack, Cordier, Spohn, Segond, Abel,
  Daviller, {\"{O}}sth, Krajnovi{\'{c}}, and Niven}]{Kaiser2014}
\bibinfo{author}{E.~Kaiser}, \bibinfo{author}{B.~Noack},
  \bibinfo{author}{L.~Cordier}, \bibinfo{author}{A.~Spohn},
  \bibinfo{author}{M.~Segond}, \bibinfo{author}{M.~Abel},
  \bibinfo{author}{G.~Daviller}, \bibinfo{author}{J.~{\"{O}}sth},
  \bibinfo{author}{S.~Krajnovi{\'{c}}}, \bibinfo{author}{R.~Niven},
\newblock \bibinfo{title}{Cluster-based reduced-order modelling of a mixing
  layer},
\newblock \bibinfo{journal}{Journal of Fluid Mechanics} \bibinfo{volume}{754}
  (\bibinfo{year}{2014}) \bibinfo{pages}{365--414}.
\bibitem[{Wang et~al.(2018)Wang, Xiao, Fang, Govindan, Pain, and
  Guo}]{Wang2018}
\bibinfo{author}{Z.~Wang}, \bibinfo{author}{D.~Xiao},
  \bibinfo{author}{F.~Fang}, \bibinfo{author}{R.~Govindan},
  \bibinfo{author}{C.~C. Pain}, \bibinfo{author}{Y.~Guo},
\newblock \bibinfo{title}{Model identification of reduced order fluid dynamics
  systems using deep learning},
\newblock \bibinfo{journal}{International Journal for Numerical Methods in
  Fluids} \bibinfo{volume}{86} (\bibinfo{year}{2018})
  \bibinfo{pages}{255--268}.
\bibitem[{{Rasheed} et~al.(2020){Rasheed}, {San}, and {Kvamsdal}}]{Rasheed2020}
\bibinfo{author}{A.~{Rasheed}}, \bibinfo{author}{O.~{San}},
  \bibinfo{author}{T.~{Kvamsdal}},
\newblock \bibinfo{title}{Digital twin: Values, challenges and enablers from a
  modeling perspective},
\newblock \bibinfo{journal}{IEEE Access} \bibinfo{volume}{8}
  (\bibinfo{year}{2020}) \bibinfo{pages}{21980--22012}.
\bibitem[{Moya et~al.(2020)Moya, Alfaro, Gonzalez, Chinesta, and
  Cueto}]{Moya2020}
\bibinfo{author}{B.~Moya}, \bibinfo{author}{I.~Alfaro},
  \bibinfo{author}{D.~Gonzalez}, \bibinfo{author}{F.~Chinesta},
  \bibinfo{author}{E.~Cueto},
\newblock \bibinfo{title}{Physically sound, self-learning digital twins for
  sloshing fluids},
\newblock \bibinfo{journal}{PLoS One} \bibinfo{volume}{15}
  (\bibinfo{year}{2020}) \bibinfo{pages}{e0234569}.
\bibitem[{Kapteyn et~al.(ming)Kapteyn, Knezevic, Huynh, Tran, and
  Willcox}]{Kapteyn2020}
\bibinfo{author}{M.~Kapteyn}, \bibinfo{author}{D.~Knezevic},
  \bibinfo{author}{D.~Huynh}, \bibinfo{author}{M.~Tran},
  \bibinfo{author}{K.~Willcox},
\newblock \bibinfo{title}{Data-driven physics-based digital twins via a library
  of component-based reduced-order models},
\newblock \bibinfo{journal}{International Journal for Numerical Methods in
  Engineering}  (\bibinfo{year}{forthcoming}).
\bibitem[{Ahmed et~al.(2019)Ahmed, Rahman, San, Rasheed, and Navon}]{Ahmed2019}
\bibinfo{author}{S.~E. Ahmed}, \bibinfo{author}{S.~M. Rahman},
  \bibinfo{author}{O.~San}, \bibinfo{author}{A.~Rasheed},
  \bibinfo{author}{I.~M. Navon},
\newblock \bibinfo{title}{Memory embedded non-intrusive reduced order modeling
  of non-ergodic flows},
\newblock \bibinfo{journal}{Physics of Fluids} \bibinfo{volume}{31}
  (\bibinfo{year}{2019}) \bibinfo{pages}{126602}.
\bibitem[{Kherad et~al.(ming)Kherad, Moayyedi, and Fotouhi}]{Kherad2020}
\bibinfo{author}{M.~Kherad}, \bibinfo{author}{M.~K. Moayyedi},
  \bibinfo{author}{F.~Fotouhi},
\newblock \bibinfo{title}{Reduced order framework for convection dominant and
  pure diffusive problems based on combination of deep long short-term memory
  and proper orthogonal decomposition/dynamic mode decomposition methods},
\newblock \bibinfo{journal}{International Journal for Numerical Methods in
  Fluids}  (\bibinfo{year}{forthcoming}).
\bibitem[{Quilodr{\'a}n-Casas et~al.(2021)Quilodr{\'a}n-Casas, Arcucci, Pain,
  and Guo}]{quilodran2021adversarially}
\bibinfo{author}{C.~Quilodr{\'a}n-Casas}, \bibinfo{author}{R.~Arcucci},
  \bibinfo{author}{C.~Pain}, \bibinfo{author}{Y.~Guo},
\newblock \bibinfo{title}{{Adversarially trained LSTMs on reduced order models
  of urban air pollution simulations}},
\newblock \bibinfo{journal}{arXiv preprint arXiv:2101.01568}
  (\bibinfo{year}{2021}).
\bibitem[{Guo and Hesthaven(2018)}]{Guo2018}
\bibinfo{author}{M.~Guo}, \bibinfo{author}{J.~S. Hesthaven},
\newblock \bibinfo{title}{Reduced order modeling for nonlinear structural
  analysis using {G}aussian process regression},
\newblock \bibinfo{journal}{Computer Methods in Applied Mechanics and
  Engineering} \bibinfo{volume}{341} (\bibinfo{year}{2018})
  \bibinfo{pages}{807--826}.
\bibitem[{Hochreiter and Schmidhuber(1997)}]{hochreiter1997long}
\bibinfo{author}{S.~Hochreiter}, \bibinfo{author}{J.~Schmidhuber},
\newblock \bibinfo{title}{{Long Short-Term Memory}},
\newblock \bibinfo{journal}{Neural computation} \bibinfo{volume}{9}
  (\bibinfo{year}{1997}) \bibinfo{pages}{1735--1780}.
\bibitem[{Xingjian et~al.(2015)Xingjian, Chen, Wang, Yeung, Wong, and
  Woo}]{xingjian2015convolutional}
\bibinfo{author}{S.~Xingjian}, \bibinfo{author}{Z.~Chen},
  \bibinfo{author}{H.~Wang}, \bibinfo{author}{D.-Y. Yeung},
  \bibinfo{author}{W.-K. Wong}, \bibinfo{author}{W.-c. Woo},
\newblock \bibinfo{title}{{Convolutional LSTM network: A machine learning
  approach for precipitation nowcasting}},
\newblock in: \bibinfo{booktitle}{{Advances in neural information processing
  systems}}, pp. \bibinfo{pages}{802--810}.
\bibitem[{Greff et~al.(2016)Greff, Srivastava, Koutn{\'\i}k, Steunebrink, and
  Schmidhuber}]{greff2016lstm}
\bibinfo{author}{K.~Greff}, \bibinfo{author}{R.~K. Srivastava},
  \bibinfo{author}{J.~Koutn{\'\i}k}, \bibinfo{author}{B.~R. Steunebrink},
  \bibinfo{author}{J.~Schmidhuber},
\newblock \bibinfo{title}{{LSTM: A search space odyssey}},
\newblock \bibinfo{journal}{IEEE transactions on neural networks and learning
  systems} \bibinfo{volume}{28} (\bibinfo{year}{2016})
  \bibinfo{pages}{2222--2232}.
\bibitem[{Schuster and Paliwal(1997)}]{schuster1997bidirectional}
\bibinfo{author}{M.~Schuster}, \bibinfo{author}{K.~K. Paliwal},
\newblock \bibinfo{title}{{Bidirectional recurrent neural networks}},
\newblock \bibinfo{journal}{IEEE transactions on Signal Processing}
  \bibinfo{volume}{45} (\bibinfo{year}{1997}) \bibinfo{pages}{2673--2681}.
\bibitem[{Cui et~al.(2018)Cui, Ke, Pu, and Wang}]{cui2018deep}
\bibinfo{author}{Z.~Cui}, \bibinfo{author}{R.~Ke}, \bibinfo{author}{Z.~Pu},
  \bibinfo{author}{Y.~Wang},
\newblock \bibinfo{title}{{Deep bidirectional and unidirectional LSTM recurrent
  neural network for network-wide traffic speed prediction}},
\newblock \bibinfo{journal}{arXiv preprint arXiv:1801.02143}
  (\bibinfo{year}{2018}).
\bibitem[{Cui et~al.(2020)Cui, Ke, Pu, and Wang}]{cui2020stacked}
\bibinfo{author}{Z.~Cui}, \bibinfo{author}{R.~Ke}, \bibinfo{author}{Z.~Pu},
  \bibinfo{author}{Y.~Wang},
\newblock \bibinfo{title}{Stacked bidirectional and unidirectional lstm
  recurrent neural network for forecasting network-wide traffic state with
  missing values},
\newblock \bibinfo{journal}{Transportation Research Part C: Emerging
  Technologies} \bibinfo{volume}{118} (\bibinfo{year}{2020})
  \bibinfo{pages}{102674}.
\bibitem[{Graves et~al.(2013)Graves, Jaitly, and Mohamed}]{graves2013hybrid}
\bibinfo{author}{A.~Graves}, \bibinfo{author}{N.~Jaitly},
  \bibinfo{author}{A.-r. Mohamed},
\newblock \bibinfo{title}{{Hybrid speech recognition with deep bidirectional
  LSTM}},
\newblock in: \bibinfo{booktitle}{{2013 IEEE workshop on automatic speech
  recognition and understanding}}, \bibinfo{organization}{IEEE}, pp.
  \bibinfo{pages}{273--278}.
\bibitem[{Liu and Guo(2019)}]{liu2019bidirectional}
\bibinfo{author}{G.~Liu}, \bibinfo{author}{J.~Guo},
\newblock \bibinfo{title}{Bidirectional {LSTM} with attention mechanism and
  convolutional layer for text classification},
\newblock \bibinfo{journal}{Neurocomputing} \bibinfo{volume}{337}
  (\bibinfo{year}{2019}) \bibinfo{pages}{325--338}.
\bibitem[{Elsheikh et~al.(2019)Elsheikh, Yacout, and
  Ouali}]{elsheikh2019bidirectional}
\bibinfo{author}{A.~Elsheikh}, \bibinfo{author}{S.~Yacout},
  \bibinfo{author}{M.-S. Ouali},
\newblock \bibinfo{title}{Bidirectional handshaking {LSTM} for remaining useful
  life prediction},
\newblock \bibinfo{journal}{Neurocomputing} \bibinfo{volume}{323}
  (\bibinfo{year}{2019}) \bibinfo{pages}{148--156}.
\bibitem[{Goodfellow et~al.(2016)Goodfellow, Bengio, Courville, and
  Bengio}]{goodfellow2016deep}
\bibinfo{author}{I.~Goodfellow}, \bibinfo{author}{Y.~Bengio},
  \bibinfo{author}{A.~Courville}, \bibinfo{author}{Y.~Bengio},
  \bibinfo{title}{Deep learning}, volume~\bibinfo{volume}{1},
  \bibinfo{publisher}{MIT press Cambridge}, \bibinfo{year}{2016}.
\bibitem[{Goodfellow et~al.(2014)Goodfellow, Pouget-Abadie, Mirza, Xu,
  Warde-Farley, Ozair, Courville, and Bengio}]{goodfellow:14}
\bibinfo{author}{I.~Goodfellow}, \bibinfo{author}{J.~Pouget-Abadie},
  \bibinfo{author}{M.~Mirza}, \bibinfo{author}{B.~Xu},
  \bibinfo{author}{D.~Warde-Farley}, \bibinfo{author}{S.~Ozair},
  \bibinfo{author}{A.~Courville}, \bibinfo{author}{Y.~Bengio},
\newblock \bibinfo{title}{{Generative adversarial nets}},
\newblock in: \bibinfo{booktitle}{{Advances in neural information processing
  systems}}, \bibinfo{year}{2014}, pp. \bibinfo{pages}{2672--2680}.
\bibitem[{Karras et~al.(2019)Karras, Laine, and Aila}]{karras2019style}
\bibinfo{author}{T.~Karras}, \bibinfo{author}{S.~Laine},
  \bibinfo{author}{T.~Aila},
\newblock \bibinfo{title}{{A Style-Based Generator Architecture for Generative
  Adversarial Networks}},
\newblock in: \bibinfo{booktitle}{Proceedings of the IEEE conference on
  computer vision and pattern recognition}, \bibinfo{year}{2019}, pp.
  \bibinfo{pages}{4401--4410}.
\bibitem[{Karras et~al.(2020)Karras, Laine, Aittala, Hellsten, Lehtinen, and
  Aila}]{karras2020analyzing}
\bibinfo{author}{T.~Karras}, \bibinfo{author}{S.~Laine},
  \bibinfo{author}{M.~Aittala}, \bibinfo{author}{J.~Hellsten},
  \bibinfo{author}{J.~Lehtinen}, \bibinfo{author}{T.~Aila},
\newblock \bibinfo{title}{{Analyzing and Improving the Image Quality of
  StyleGAN}},
\newblock in: \bibinfo{booktitle}{Proceedings of the IEEE/CVF Conference on
  Computer Vision and Pattern Recognition}, pp. \bibinfo{pages}{8110--8119}.
\bibitem[{Chu et~al.(2017)Chu, Zhmoginov, and Sandler}]{chu2017cyclegan}
\bibinfo{author}{C.~Chu}, \bibinfo{author}{A.~Zhmoginov},
  \bibinfo{author}{M.~Sandler},
\newblock \bibinfo{title}{{CycleGAN, a Master of Steganography}},
\newblock \bibinfo{journal}{arXiv preprint arXiv:1712.02950}
  (\bibinfo{year}{2017}).
\bibitem[{Frid-Adar et~al.(2018)Frid-Adar, Diamant, Klang, Amitai, Goldberger,
  and Greenspan}]{frid2018gan}
\bibinfo{author}{M.~Frid-Adar}, \bibinfo{author}{I.~Diamant},
  \bibinfo{author}{E.~Klang}, \bibinfo{author}{M.~Amitai},
  \bibinfo{author}{J.~Goldberger}, \bibinfo{author}{H.~Greenspan},
\newblock \bibinfo{title}{{GAN-based synthetic medical image augmentation for
  increased CNN performance in liver lesion classification}},
\newblock \bibinfo{journal}{Neurocomputing} \bibinfo{volume}{321}
  (\bibinfo{year}{2018}) \bibinfo{pages}{321--331}.
\bibitem[{Liu et~al.(2018)Liu, Qin, Wan, and Luo}]{liu2018auto}
\bibinfo{author}{Y.~Liu}, \bibinfo{author}{Z.~Qin}, \bibinfo{author}{T.~Wan},
  \bibinfo{author}{Z.~Luo},
\newblock \bibinfo{title}{Auto-painter: Cartoon image generation from sketch by
  using conditional {W}asserstein generative adversarial networks},
\newblock \bibinfo{journal}{Neurocomputing} \bibinfo{volume}{311}
  (\bibinfo{year}{2018}) \bibinfo{pages}{78--87}.
\bibitem[{Silva et~al.(2021)Silva, Heaney, Li, and Pain}]{Silva2020}
\bibinfo{author}{V.~L.~S. Silva}, \bibinfo{author}{C.~E. Heaney},
  \bibinfo{author}{Y.~Li}, \bibinfo{author}{C.~C. Pain}, \bibinfo{title}{{Data
  Assimilation Predictive GAN (DA-PredGAN): applied to determine the spread of
  COVID-19}}, \bibinfo{year}{2021}. \bibinfo{note}{In preparation}.
\bibitem[{Yang et~al.(2020)Yang, Zeng, Wang, Wong, Liang, Zanin, Liu, Cao, Gao,
  Mai et~al.}]{yang2020modified}
\bibinfo{author}{Z.~Yang}, \bibinfo{author}{Z.~Zeng},
  \bibinfo{author}{K.~Wang}, \bibinfo{author}{S.-S. Wong},
  \bibinfo{author}{W.~Liang}, \bibinfo{author}{M.~Zanin},
  \bibinfo{author}{P.~Liu}, \bibinfo{author}{X.~Cao}, \bibinfo{author}{Z.~Gao},
  \bibinfo{author}{Z.~Mai}, et~al.,
\newblock \bibinfo{title}{{Modified SEIR and AI prediction of the epidemics
  trend of COVID-19 in China under public health interventions}},
\newblock \bibinfo{journal}{Journal of Thoracic Disease} \bibinfo{volume}{12}
  (\bibinfo{year}{2020}) \bibinfo{pages}{165}.
\bibitem[{Ardabili et~al.(2020)Ardabili, Mosavi, Ghamisi, Ferdinand,
  Varkonyi-Koczy, Reuter, Rabczuk, and Atkinson}]{ardabili2020covid}
\bibinfo{author}{S.~F. Ardabili}, \bibinfo{author}{A.~Mosavi},
  \bibinfo{author}{P.~Ghamisi}, \bibinfo{author}{F.~Ferdinand},
  \bibinfo{author}{A.~R. Varkonyi-Koczy}, \bibinfo{author}{U.~Reuter},
  \bibinfo{author}{T.~Rabczuk}, \bibinfo{author}{P.~M. Atkinson},
\newblock \bibinfo{title}{{COVID-19 Outbreak Prediction with Machine
  Learning}},
\newblock \bibinfo{journal}{Available at SSRN 3580188}  (\bibinfo{year}{2020}).
\bibitem[{Chimmula and Zhang(2020)}]{chimmula2020time}
\bibinfo{author}{V.~K.~R. Chimmula}, \bibinfo{author}{L.~Zhang},
\newblock \bibinfo{title}{{Time series forecasting of COVID-19 transmission in
  Canada using LSTM networks}},
\newblock \bibinfo{journal}{Chaos, Solitons \& Fractals}
  (\bibinfo{year}{2020}) \bibinfo{pages}{109864}.
\bibitem[{Ayyoubzadeh et~al.(2020)Ayyoubzadeh, Ayyoubzadeh, Zahedi, Ahmadi, and
  Kalhori}]{ayyoubzadeh2020predicting}
\bibinfo{author}{S.~M. Ayyoubzadeh}, \bibinfo{author}{S.~M. Ayyoubzadeh},
  \bibinfo{author}{H.~Zahedi}, \bibinfo{author}{M.~Ahmadi},
  \bibinfo{author}{S.~R.~N. Kalhori},
\newblock \bibinfo{title}{{Predicting COVID-19 Incidence Through Analysis of
  Google Trends Data in Iran: Data Mining and Deep Learning Pilot Study}},
\newblock \bibinfo{journal}{JMIR Public Health and Surveillance}
  \bibinfo{volume}{6} (\bibinfo{year}{2020}) \bibinfo{pages}{e18828}.
\bibitem[{Khalifa et~al.(2020)Khalifa, Taha, Hassanien, and
  Elghamrawy}]{khalifa2020detection}
\bibinfo{author}{N.~E.~M. Khalifa}, \bibinfo{author}{M.~H.~N. Taha},
  \bibinfo{author}{A.~E. Hassanien}, \bibinfo{author}{S.~Elghamrawy},
\newblock \bibinfo{title}{{Detection of coronavirus (COVID-19) associated
  pneumonia based on generative adversarial networks and a fine-tuned deep
  transfer learning model using chest X-ray dataset}},
\newblock \bibinfo{journal}{arXiv preprint arXiv:2004.01184}
  (\bibinfo{year}{2020}).
\bibitem[{Wang and Wong(2020)}]{wang2020covid}
\bibinfo{author}{L.~Wang}, \bibinfo{author}{A.~Wong},
\newblock \bibinfo{title}{{COVID-Net: A Tailored Deep Convolutional Neural
  Network Design for Detection of COVID-19 Cases from Chest X-Ray Images}},
\newblock \bibinfo{journal}{arXiv preprint arXiv:2003.09871}
  (\bibinfo{year}{2020}).
\bibitem[{Wang et~al.(2020)Wang, Yang, Li, Nadler, Arcucci, Huang, Teng, and
  Guo}]{wang2020bayesian}
\bibinfo{author}{S.~Wang}, \bibinfo{author}{X.~Yang}, \bibinfo{author}{L.~Li},
  \bibinfo{author}{P.~Nadler}, \bibinfo{author}{R.~Arcucci},
  \bibinfo{author}{Y.~Huang}, \bibinfo{author}{Z.~Teng},
  \bibinfo{author}{Y.~Guo},
\newblock \bibinfo{title}{{A Bayesian Updating Scheme for Pandemics: Estimating
  the Infection Dynamics of COVID-19}},
\newblock \bibinfo{journal}{IEEE Computational Intelligence Magazine}
  \bibinfo{volume}{15} (\bibinfo{year}{2020}) \bibinfo{pages}{23--33}.
\bibitem[{Bao et~al.(2020)Bao, Zhou, Zhang, Li, and Xie}]{bao2020covid}
\bibinfo{author}{H.~Bao}, \bibinfo{author}{X.~Zhou},
  \bibinfo{author}{Y.~Zhang}, \bibinfo{author}{Y.~Li},
  \bibinfo{author}{Y.~Xie},
\newblock \bibinfo{title}{{COVID-GAN: Estimating Human Mobility Responses to
  COVID-19 Pandemic through Spatio-Temporal Conditional Generative Adversarial
  Networks}},
\newblock in: \bibinfo{booktitle}{Proceedings of the 28th International
  Conference on Advances in Geographic Information Systems}, pp.
  \bibinfo{pages}{273--282}.
\bibitem[{Yoon et~al.(2019)Yoon, Jarrett, and van~der Schaar}]{yoon2019time}
\bibinfo{author}{J.~Yoon}, \bibinfo{author}{D.~Jarrett},
  \bibinfo{author}{M.~van~der Schaar},
\newblock \bibinfo{title}{Time-series generative adversarial networks}
  (\bibinfo{year}{2019}).
\bibitem[{Breuel(2015)}]{breuel2015benchmarking}
\bibinfo{author}{T.~M. Breuel},
\newblock \bibinfo{title}{Benchmarking of lstm networks},
\newblock \bibinfo{journal}{arXiv preprint arXiv:1508.02774}
  (\bibinfo{year}{2015}).
\bibitem[{{Institute for Disease Modelling}(2020)}]{idm:2020}
\bibinfo{author}{{Institute for Disease Modelling}}, \bibinfo{title}{{SEIR and
  SEIRS models}}, \bibinfo{year}{2020}.
\bibitem[{Nadler et~al.(2020)Nadler, Wang, Arcucci, Yang, and
  Guo}]{nadler2020epidemiological}
\bibinfo{author}{P.~Nadler}, \bibinfo{author}{S.~Wang},
  \bibinfo{author}{R.~Arcucci}, \bibinfo{author}{X.~Yang},
  \bibinfo{author}{Y.~Guo},
\newblock \bibinfo{title}{An epidemiological modelling approach for {COVID-19}
  via data assimilation},
\newblock \bibinfo{journal}{European Journal of Epidemiology}
  \bibinfo{volume}{35} (\bibinfo{year}{2020}) \bibinfo{pages}{749--761}.
\bibitem[{{UK Government}(2020)}]{govuk2020}
\bibinfo{author}{{UK Government}}, \bibinfo{title}{{COVID-19: infection
  prevention and control (IPC)}}, \bibinfo{year}{2020}.
\bibitem[{Wengert(1964)}]{wengert:64}
\bibinfo{author}{R.~E. Wengert},
\newblock \bibinfo{title}{{A simple automatic derivative evaluation program}},
\newblock \bibinfo{journal}{Communications of the ACM} \bibinfo{volume}{7}
  (\bibinfo{year}{1964}) \bibinfo{pages}{463--464}.
\bibitem[{Linnainmaa(1976)}]{linnainmaa:76}
\bibinfo{author}{S.~Linnainmaa},
\newblock \bibinfo{title}{{Taylor expansion of the accumulated rounding
  error}},
\newblock \bibinfo{journal}{BIT Numerical Mathematics} \bibinfo{volume}{16}
  (\bibinfo{year}{1976}) \bibinfo{pages}{146--160}.
\bibitem[{Baydin et~al.(2017)Baydin, Pearlmutter, Radul, and
  Siskind}]{baydin:17}
\bibinfo{author}{A.~G. Baydin}, \bibinfo{author}{B.~A. Pearlmutter},
  \bibinfo{author}{A.~A. Radul}, \bibinfo{author}{J.~M. Siskind},
\newblock \bibinfo{title}{{Automatic differentiation in machine learning: a
  survey}},
\newblock \bibinfo{journal}{The Journal of Machine Learning Research}
  \bibinfo{volume}{18} (\bibinfo{year}{2017}) \bibinfo{pages}{5595--5637}.
\bibitem[{Abadi et~al.(2015)Abadi, Agarwal, Barham, Brevdo, Chen, Citro,
  Corrado, Davis, Dean, Devin, Ghemawat, Goodfellow, Harp, Irving, Isard, Jia,
  Jozefowicz, Kaiser, Kudlur, Levenberg, Man\'{e}, Monga, Moore, Murray, Olah,
  Schuster, Shlens, Steiner, Sutskever, Talwar, Tucker, Vanhoucke, Vasudevan,
  Vi\'{e}gas, Vinyals, Warden, Wattenberg, Wicke, Yu, and
  Zheng}]{tensorflow:2015}
\bibinfo{author}{M.~Abadi}, \bibinfo{author}{A.~Agarwal},
  \bibinfo{author}{P.~Barham}, \bibinfo{author}{E.~Brevdo},
  \bibinfo{author}{Z.~Chen}, \bibinfo{author}{C.~Citro}, \bibinfo{author}{G.~S.
  Corrado}, \bibinfo{author}{A.~Davis}, \bibinfo{author}{J.~Dean},
  \bibinfo{author}{M.~Devin}, \bibinfo{author}{S.~Ghemawat},
  \bibinfo{author}{I.~Goodfellow}, \bibinfo{author}{A.~Harp},
  \bibinfo{author}{G.~Irving}, \bibinfo{author}{M.~Isard},
  \bibinfo{author}{Y.~Jia}, \bibinfo{author}{R.~Jozefowicz},
  \bibinfo{author}{L.~Kaiser}, \bibinfo{author}{M.~Kudlur},
  \bibinfo{author}{J.~Levenberg}, \bibinfo{author}{D.~Man\'{e}},
  \bibinfo{author}{R.~Monga}, \bibinfo{author}{S.~Moore},
  \bibinfo{author}{D.~Murray}, \bibinfo{author}{C.~Olah},
  \bibinfo{author}{M.~Schuster}, \bibinfo{author}{J.~Shlens},
  \bibinfo{author}{B.~Steiner}, \bibinfo{author}{I.~Sutskever},
  \bibinfo{author}{K.~Talwar}, \bibinfo{author}{P.~Tucker},
  \bibinfo{author}{V.~Vanhoucke}, \bibinfo{author}{V.~Vasudevan},
  \bibinfo{author}{F.~Vi\'{e}gas}, \bibinfo{author}{O.~Vinyals},
  \bibinfo{author}{P.~Warden}, \bibinfo{author}{M.~Wattenberg},
  \bibinfo{author}{M.~Wicke}, \bibinfo{author}{Y.~Yu},
  \bibinfo{author}{X.~Zheng}, \bibinfo{title}{{ {TensorFlow}: Large-Scale
  Machine Learning on Heterogeneous Systems}}, \bibinfo{year}{2015}.
  \bibinfo{note}{Software available from tensorflow.org}.
\bibitem[{Chollet et~al.(2015)}]{chollet2015keras}
\bibinfo{author}{F.~Chollet}, et~al., \bibinfo{title}{Keras},
  \bibinfo{howpublished}{\url{https://keras.io}}, \bibinfo{year}{2015}.
\bibitem[{Radford et~al.(2015)Radford, Metz, and Chintala}]{radford:15}
\bibinfo{author}{A.~Radford}, \bibinfo{author}{L.~Metz},
  \bibinfo{author}{S.~Chintala},
\newblock \bibinfo{title}{{Unsupervised Representation Learning with Deep
  Convolutional Generative Adversarial Networks}},
\newblock \bibinfo{journal}{arXiv preprint arXiv:1511.06434}
  (\bibinfo{year}{2015}).
\bibitem[{Zhu et~al.(2019)Zhu, Ye, Fu, Liu, and
  Shen}]{zhu2019electrocardiogram}
\bibinfo{author}{F.~Zhu}, \bibinfo{author}{F.~Ye}, \bibinfo{author}{Y.~Fu},
  \bibinfo{author}{Q.~Liu}, \bibinfo{author}{B.~Shen},
\newblock \bibinfo{title}{{Electrocardiogram generation with a bidirectional
  LSTM-CNN generative adversarial network}},
\newblock \bibinfo{journal}{Scientific reports} \bibinfo{volume}{9}
  (\bibinfo{year}{2019}) \bibinfo{pages}{1--11}.
\bibitem[{Mogren(2016)}]{mogren2016c}
\bibinfo{author}{O.~Mogren},
\newblock \bibinfo{title}{{C-RNN-GAN: Continuous recurrent neural networks with
  adversarial training}},
\newblock \bibinfo{journal}{arXiv preprint arXiv:1611.09904}
  (\bibinfo{year}{2016}).
\bibitem[{Shahid et~al.(2020)Shahid, Zameer, and
  Muneeb}]{shahid2020predictions}
\bibinfo{author}{F.~Shahid}, \bibinfo{author}{A.~Zameer},
  \bibinfo{author}{M.~Muneeb},
\newblock \bibinfo{title}{Predictions for covid-19 with deep learning models of
  lstm, gru and bi-lstm},
\newblock \bibinfo{journal}{Chaos, Solitons \& Fractals} \bibinfo{volume}{140}
  (\bibinfo{year}{2020}) \bibinfo{pages}{110212}.
\bibitem[{Chatterjee et~al.(2020)Chatterjee, Gerdes, and
  Martinez}]{chatterjee2020statistical}
\bibinfo{author}{A.~Chatterjee}, \bibinfo{author}{M.~W. Gerdes},
  \bibinfo{author}{S.~G. Martinez},
\newblock \bibinfo{title}{Statistical explorations and univariate timeseries
  analysis on covid-19 datasets to understand the trend of disease spreading
  and death},
\newblock \bibinfo{journal}{Sensors} \bibinfo{volume}{20}
  (\bibinfo{year}{2020}) \bibinfo{pages}{3089}.

\end{thebibliography}







\end{document}